\documentclass[10pt,twocolumn,letterpaper]{article}

\usepackage{iccv}
\usepackage{times}
\usepackage{graphicx}
\usepackage{epsfig}
\usepackage{amsmath}
\usepackage{amssymb}
\usepackage{multirow}
\usepackage{subfigure}
\usepackage{booktabs}

\usepackage[pagebackref=true,breaklinks=true,letterpaper=true,colorlinks,bookmarks=false]{hyperref}

\iccvfinalcopy 

\author{Xinzhu Ma\textsuperscript{1}, \ \  
	Zhihui Wang\textsuperscript{1, 2}, \ \  
	Haojie Li\textsuperscript{1, 2, *},  \ \ 
	Pengbo Zhang\textsuperscript{1}, \ \ 
	Xin Fan\textsuperscript{1, 2}, \ \
	Wanli Ouyang\textsuperscript{3} \\  
\textsuperscript{1}Dalian University of Technology, China \\
\textsuperscript{3}Key Laboratory for Ubiquitous Network and Service Software of Liaoning Province, China\\
\textsuperscript{4}The University of Sydney, SenseTime Computer Vision Research Group, Australia\\
{\tt\small \{maxinzhu@mail., zhwang@, hjli@, bobo96@mail., xin.fan\}dlut.edu.cn}
\\
{\tt\small wanli.ouyang@sydney.edu.au}
}

\ificcvfinal\fi
\begin{document}

\title{Accurate Monocular Object Detection via Color-Embedded 3D Reconstruction for Autonomous Driving}

\maketitle

\begin{abstract}
In this paper, we propose a monocular 3D object detection framework in the domain of autonomous driving. 
Unlike previous image-based methods which focus on RGB feature extracted from 2D images, our method solves this problem in the reconstructed 3D space in order to exploit 3D contexts explicitly. 
To this end, we first leverage a stand-alone module to transform the input data from 2D image plane to 3D point clouds space for a better input representation, then we perform the 3D detection using PointNet backbone net to obtain objects' 3D locations, dimensions and orientations. 
To enhance the discriminative capability of point clouds, we propose a multi-modal features fusion module to embed the complementary RGB cue into the generated point clouds representation. 
We argue that it is more effective to infer the 3D bounding boxes from the generated 3D scene space (i.e., X,Y, Z space) compared to the image plane (i.e., R,G,B image plane).
Evaluation on the challenging KITTI dataset shows that our approach boosts the performance of state-of-the-art monocular approach by a large margin.

\noindent
\end{abstract}

\vspace{-5pt}
\section{Introduction}
\label{sec:one}

\renewcommand{\thefootnote}{}
\footnote{* Corresponding author: hjli@dlut.edu.cn}
\vspace{-13pt}

In recent years, with the development of technologies in computer vision and deep learning \cite{he2016deep, simonyan2014very, sun2018fishnet}, numerous impressive methods are proposed for accurate 2D object detection \cite{girshick2014rich, girshick2015fast, he2017mask, ren2015faster, lin2018focal, Ouyang_2017_ICCV, zeng2017crafting, liu2018deep}.
However, beyond getting 2D bounding boxes or pixel masks, 3D object detection is eagerly in demand in many applications such as autonomous driving and robotic applications because it can describe objects in a more realistic way.
Now, this problem received more and more the concern of scholars.
Because LiDAR provides reliable depth information that can be used to accurately localize objects and characterize their shapes, many approaches \cite{li20173d, luo2018fast, maturana2015voxnet, qi2017frustum, chen2017multi, ren2016three, yang2018pixor} use LiDAR point clouds as their input, and get impressive detection results in autonomous
driving scenarios. 
In contrast, some other studies \cite{chabot2017deep, chen20153d, chen2016monocular, xu2018multi, mousavian20173d, song2014sliding, Li_2019_CVPR} are devoted to replace the LiDAR with cheaper monocular cameras, which are readily available in daily life.
As LiDAR is much more expensive and inspired by the remarkable progress in image-based depth prediction techniques, this paper focuses on the high performance detection of 3D object utilizing only monocular images.
However, image-based 3D detection is very challenging, and there is a huge gap between the performance of image-based methods and LiDAR-based methods.
We show in this work that we can largely boost the performance of image-based 3D detection by transforming the input data representation.

\begin{figure}[t]
\begin{center}
{\includegraphics[width=0.235\textwidth, height=0.1\textwidth]{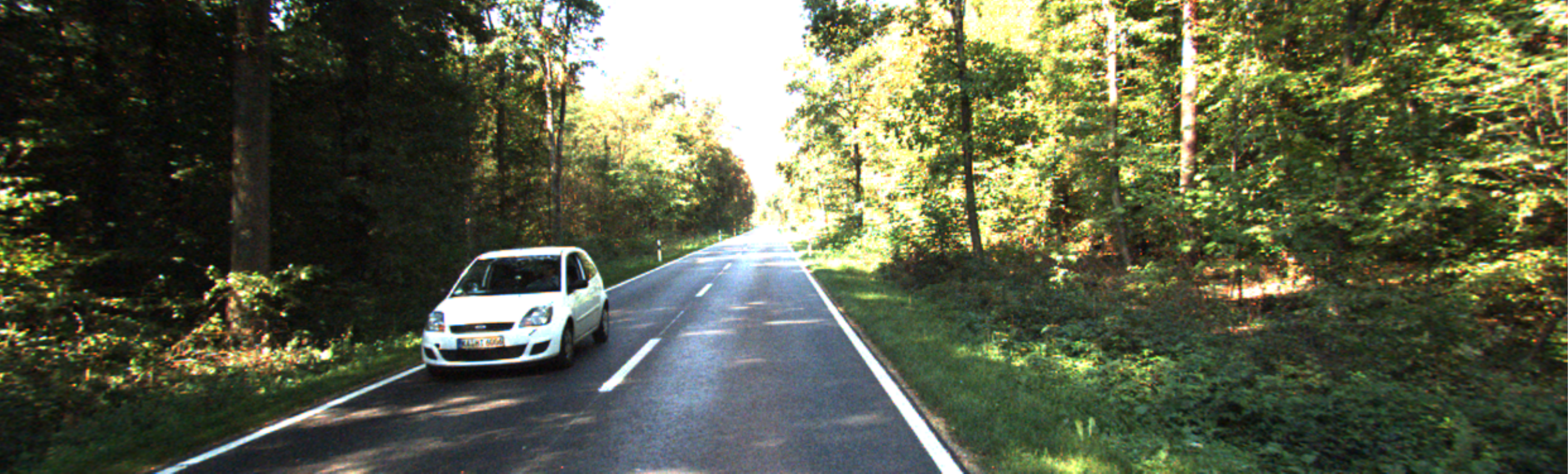}}
{\includegraphics[width=0.235\textwidth, height=0.1\textwidth]{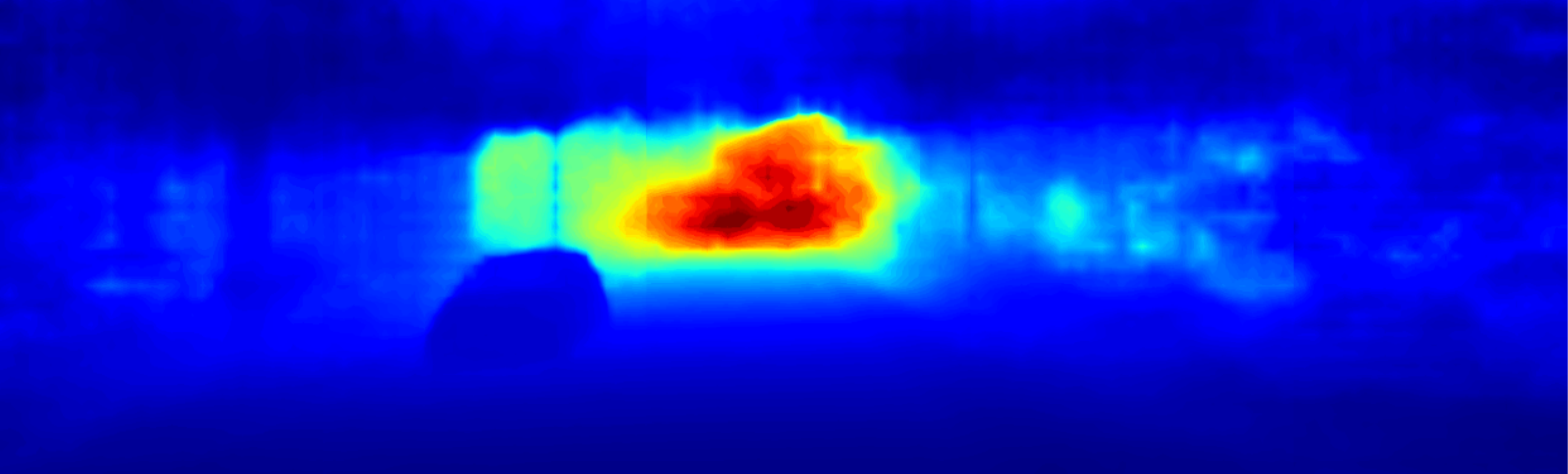}}
{\includegraphics[width=0.235\textwidth, height=0.1\textwidth]{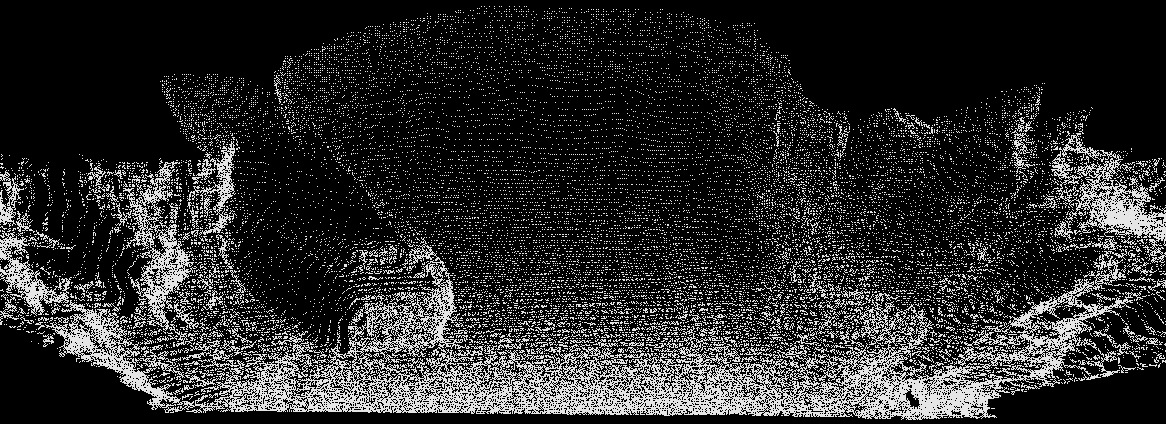}}
{\includegraphics[width=0.235\textwidth, height=0.1\textwidth]{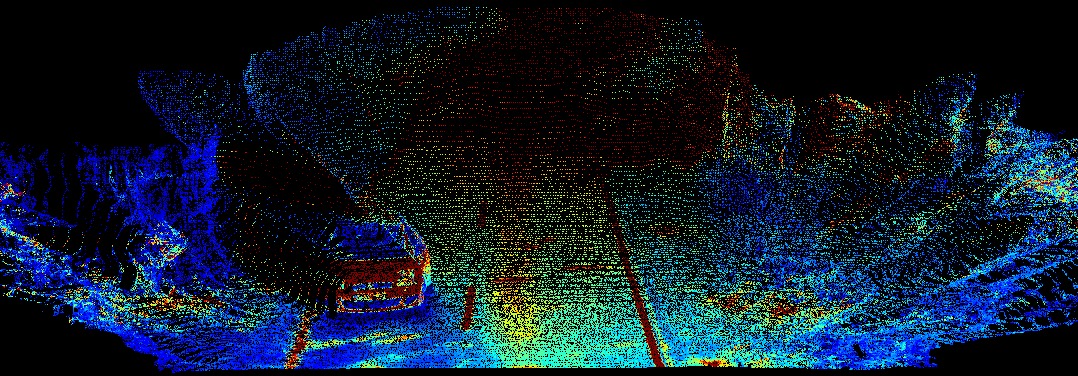}}
\end{center}
\caption{{\bf Different representations of input data.}
{\it Top left:} RGB image.
{\it Top right:} Depth map.
{\it Bottom left:} Point cloud.
{\it Bottom right:} RGB augmenting point cloud (only R-channel is mapped for this visualization).
Note that all the representations we mentioned can be generated by a single RGB image.
}
\label{fig:representation}
\end{figure}

Typical image-based 3D object detection approaches \cite{chabot2017deep, chen2016monocular, chen20153d, song2014sliding} adopted the pipeline similar to 2D detectors and mainly focused on RGB features extracted from 2D images.
However, these features are not suitable for 3D related tasks because of the lack of spatial information.
This is one of the main reasons why early studies failed to get better performance.
An intuitive solution is that we can use a CNN to predict the depth maps \cite{xu2018pad, xu2018monocular, fu2018deep} and then use them as input if we do not have the depth data available.
Although depth information is helpful to 3D scene understanding, simply using it as an additional channel of RGB images such as \cite{xu2018multi} does not compensate for the performance difference between image-based methods and LiDAR-based method.
There is no doubt that LiDAR data is much more accurate than estimated depth, here we argue that the performance gap not only due to the accuracy of the data, but also its representation (see Fig.~\ref{fig:representation} for different input representations on monocular 3D detection task).
In order to narrow the gap and and make the estimated depth a bigger role, we need a more explicit representation form such as point cloud which describes a real world 3D coordinates rather than depth with a relative position in images. 
For example, objects with different positions in 3D world may have the same coordinates in image plane, which brings difficulties for the network to estimate the final results.
The benefits for transform depth map into point cloud can be enumerated as follow: 
(1) Point cloud data shows the spatial information explicitly, which make it easier for network to learn the non-linear mapping from input to output.
(2) Richer features can be learnt by the network because some specific spatial structures exist only in 3D space.
(3) The recent significant progress of deep learning on point clouds provides a solid building brick, which we can estimate 3D detection results in a more effective and efficient way.
 
\begin{figure*}[!t]
\begin{center}
{\includegraphics[width=0.99\textwidth]{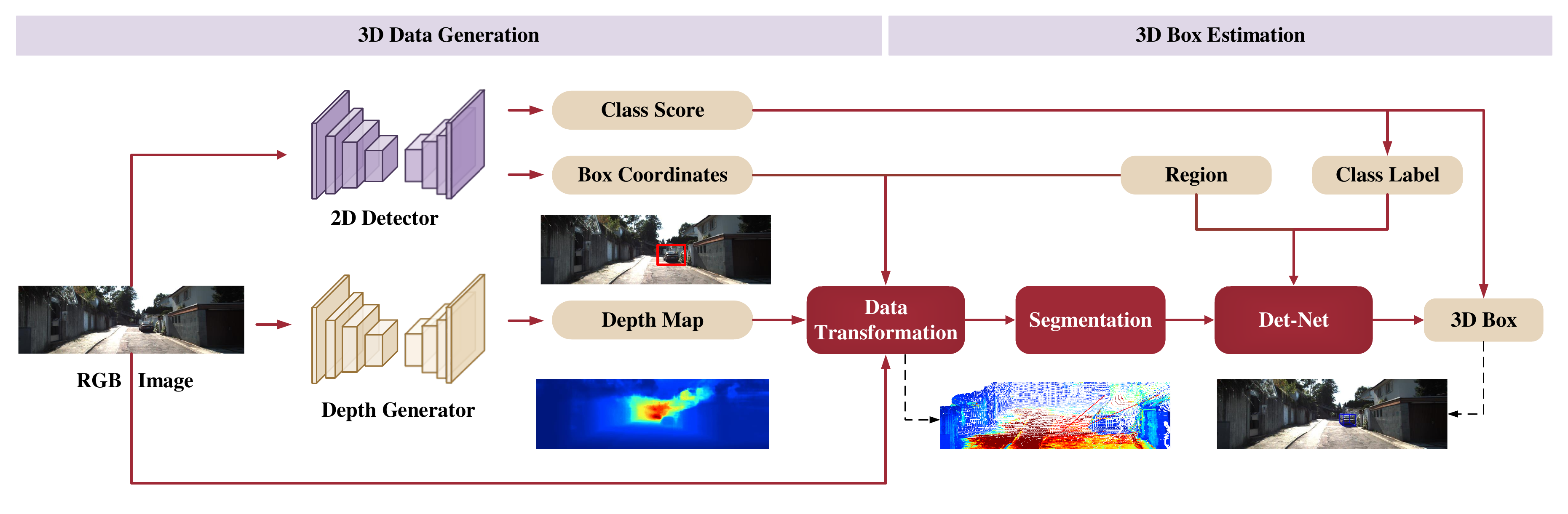}}
\end{center}
\vspace{-4pt}
\caption{The proposed framework for monocular 3D object detection.
}
\vspace{-4pt}
\label{fig:pipeline}
\end{figure*}

Based on the observations above, a monocular 3D object detection framework is proposed.
The main idea for the design of our method is to find a better input representation.
Specifically, we first learn to use front-end deep CNNs and the input RGB data to produce two intermediate tasks involving 2D detection \cite{ouyang2013joint, ouyang2017jointly, girshick2015fast} and depth estimation \cite{fu2018deep, xu2018monocular} (see Fig.~\ref{fig:pipeline}).
Then, we transform depth maps into point clouds with the help of camera calibration files in order to give the 3D information explicitly and used them as input data for subsequent steps.
Besides, another crucial component that ensures the performance of proposed method is multi-modal features fusion module.
After aggregating RGB information which is complementary to 3D point clouds, the discriminative capability of features used to describe 3D object are further enhanced.
Note that, when the optimization of the all networks are finished, the inference phase is only based on the RGB input.

The contributions of this paper can be summarized as:
\begin{itemize}
\item 
We propose a new framework for monocular 3D object detection which transforms 2D image to 3D point cloud and performs the 3D detection effectively and efficiently.
\item
We design an features fusion strategy to fully exploit the advantages of RGB cue and point cloud to boost the detection performance, which can be also applied in other scenarios such as LiDAR-based 3D detection.
\item
Evaluation on the challenging KITTI dataset~\cite{Geiger2012CVPR} shows our method outperform all state-of-the-art monocular methods by around 15\% and 11\% higher AP on 3D localization and detection tasks, respectively.
\end{itemize}

\section{Related Work}
\label{sec:two}

We briefly review existing works on 3D object detection task based on LiDAR and images in autonomous driving scenario.

\noindent
{\bf Image-based 3D Object Detection:}
In the early works, monocular-based methods share similar framework with 2D detection~\cite{girshick2015fast}, but it is much more complicated for estimating the 3D coordinates (x, y, z) of object center, since only image appearance cannot decide the absolute physical location.
Mono3D~\cite{chen2016monocular} and 3DOP~\cite{chen20153d} focus on 3D object proposals generation using prior knowledge (e.g., object size, ground plane) from monocular and stereo images, respectively.
Deep3DBox~\cite{mousavian20173d} introduces geometric constraints based on the fact that the 3D bounding box should fit tightly into 2D detection bounding box.
Deep MANTA \cite{chabot2017deep} encodes 3D vehicle information using key points, since they are rigid objects with well known geometry.
Then the vehicle recognition in Deep MANTA can be considered as extra key points detection.

Although these methods propose some effective prior knowledge or reasonable constraints, they fail to get promising performance because of the lack of spatial information.
Another recently proposed method \cite{xu2018multi} for monocular 3D object detection introduces a multi-level fusion based scheme utilizes a stand-alone module to estimate the disparity information and fuse it with RGB information in the input data encoding, 2D box estimation and 3D box estimation phase, respectively.
Although it used depth (or disparity) many times, they only regard it as auxiliary information of RGB features, and do not make full use of its potential value.     
In comparison, our method takes the generated depth as the core feature and transform it into 3D space to explicitly make use of its spatial information.
Pseudo-LiDAR~\cite{wang2019pseudo} also find that data presentation plays an important role in 3D detection task.
It pays more attention to verify the universality of point cloud representation and applies the generated points to some different existing 3D detection methods {\it without any modifications}.
In contrast, in addition to transforming data representations, we further design a dedicated 3D detection framework for monocular images.

\noindent
{\bf LiDAR-based 3D Object Detection:}
Although our approach is for monocular image data, we transform the data representation into point cloud which is same to LiDAR-based methods.
So, we also introduce some typical approach based on LiDAR.
MV3D \cite{chen2017multi} encode 3D point clouds with multi-view feature maps, enabling region-based representation for multimodal fusion.
With the development of deep learning on raw point clouds \cite{qi2017pointnet, qi2017pointnet++, huang2018recurrent}, several detection approaches only based on raw LiDAR data are also proposed.
Qi \etal \cite{qi2017frustum} extend PointNet to 3D detection task by extracting the frustum point clouds corresponding to their 2D detections.
VoxelNet \cite{zhou2017voxelnet} divides point clouds into equally spaced 3D voxels and transforms a group of points within each voxel into a unified feature representation.
Finally, 2D convolution layers are used on these high-level voxel-wise features to get spatial features and give prediction results.
Despite these two methods get a promising detection results, they do not make a good use of RGB information.
In comparison, we also introduce a RGB features fusion module to enhance the discriminative capability of point clouds.

\section{Proposed Method}
\label{sec:three}

In this section, we describe the proposed framework for monocular-based 3D object detection. 
We first present an overview of the proposed method, and then introduce the details of it. 
Finally, we show the optimization and implementation details for the overall network.

\subsection{Approach Overview}
\label{sec:3-1}

As shown in Fig. \ref{fig:pipeline}, the proposed 3D detection framework consists of two main stages.
In 3D data generation phase, we trained two deep CNNs to do intermediate tasks (2D detection and depth estimation) to get position and depth information.
In particular, we transfer the generated depth into point cloud which is a better representation for 3D detection, and then we use 2D bounding box to get the prior information about the location of the RoI (region of interest).
Finally, we extract the points in each RoI as our input data for subsequent steps.
In 3D box estimation phase, in order to improve the final task, we design two modules for background points segmentation and RGB information aggregation, respectively.
After that, we use PointNet as our backbone net to predict the 3D location, dimension and orientation for each RoI.
Note that the confidence scores of 2D boxes are assigned to their corresponding 3D boxes.

\subsection{3D Data Generation}
\noindent
\textbf{Intermediate tasks.}
As we all know that 3D detection using only monocular images is a very challenging task because image appearance can not determine the 3D coordinates of the object.
Therefore, we train two deep CNN to generate depth map and 2D bounding box to provide spatial information and position prior.
We adopt some existing algorithms to do these intermediate tasks, and give a detailed analysis of the impact of these algorithms on overall performance in experiment part.

\noindent
\textbf{Input representation.}
This work focuses more on how to use depth information than on how to get them.
We believe that one of the main reasons why previous images-based 3D detectors fails to get better results is they  don't make good use of depth maps.
Simply using depth map as an additional channel of RGB image such as \cite{xu2018monocular, manhardt2018roi}, and then expecting neural network to extract effective features automatically is not the best solution.
In contrast, we transform the estimated depth into point cloud with the help of camera calibration file provided by KITTI (see Fig.~\ref{fig:representation} for different input representations) and then use it as our data input form.
Specifically, given a pixel coordinate $(u, v)$ with depth $d$ in the 2D image space, the 3D coordinates $(x, y, z)$ in camera coordinate system can be computed as:
\begin{equation}
\label{2dto3d}
\left\{
      \begin{array}{lr}
      \vspace{4pt}
      z = d ,\\
      \vspace{4pt}
      x = (u-C_{x}) * z / f  ,\\
      y = (v-C_{y}) * z / f  ,\\
      \end{array}
\right.
\end{equation}
where $f$ is the focal length of the camera, $(C_{x}, C_{y})$ is the principal point.
The input point cloud $S$ can be generated using depth map and 2D bounding box {\bf B} as follow:
\begin{equation}
\label{pointset}
	S = \{\ p  \ | \ p \leftarrow \ {\bf F}(v), \ v\in {\bf B} \} ,
\end{equation}
where $v$ is the pixel in depth map and {\bf F} is the transforming function introduced by Eq.~\ref{2dto3d}.
It should be noted that, like most of monocular-based methods, we use camera calibration file in our approach.
Actually, we can also use a point cloud encoder-decoder net to learn a mapping from $(u, v, d)$ to $(x, y, z)$, thus we don't need camera during the testing phase any more.
In our measurements, we observe that there is no visible performance difference between these two methods.
This is because the error introduced in the point cloud generation phase is much less than the noise contained in the depth map itself.

\subsection{3D Box Estimation}

\noindent
\textbf{Point segmentation.}
After the 3D data generation phase, the input data is encoded as points cloud. 
However, there are many background points in these data and these background points should be discarded in order to estimate the position of target accurately.
Qi \etal~\cite{qi2017frustum} propose a 3D instance segmentation PointNet to solve this problem in LiDAR data.
But that strategy requires additional pre-processing to generate segmentation labels from 3D object ground truth.
More importantly, there will be severe noise even if we use the same labelling method because the points we reconstruct are relatively unstable.
For these reasons, we propose a simple but effective segmentation method based on depth prior to segment the points.
Specifically, we first compute the depth mean in each 2D bounding box in order to get the approximate position of RoI, and use it as the threshold.
All points with Z-channel value greater than this threshold are considered as background points.
The processed point set $S'$ can be expressed as:

\begin{equation}
	S' = \{\ p  \ | \ p_{v} \leq \frac{\sum_{p\in S}p_{v} }{|S|} + r, \ p\in S \} ,
\end{equation}
where $p_{v}$ denotes the Z-channel value (which is equal to depth) of the point and $r$ is a bias used to correct the threshold.
Finally, we randomly select a fixed number of points in point set $S'$ as the output of this module in order to ensuring consistency of number of subsequent network's input points.

\noindent
\textbf{3D box estimation.}
Before we estimate final 3D results, we follow \cite{qi2017frustum} to predict the center $\delta$ of RoI using a lightweight network and use it to update the point cloud as follow:
\begin{equation}
	S'' = \{\ p  \ | \ p - \delta , \ p\in S' \} ,
\end{equation}
where $S''$ is the set of points we used to do final task.
Then, we choose PointNet~\cite{qi2017pointnet} as our 3D detection backbone network to estimate the 3D object which is encoded by its center $(x, y, z)$, size $(h, w, l)$ and heading angle $\theta$.
Same as other works, we only consider one orientation because of the assumption that the road surface is flat and the other two angles do not have possible variation.
One other thing to note is that the center $C$ we estimate here is a 'residual' center, which means the real center is $C+\delta$.
Finally, we assign the confidence scores of the 2D bounding boxes to their corresponding 3D detection results.

\begin{table*}[!ht]
\begin{center}
\begin{tabular}{c||c|c|c|c|c|c|c}
\hline
\multirow{2}{*}{Method} & \multirow{2}{*}{Data} & \multicolumn{3}{c|}{IoU=0.5} & \multicolumn{3}{c}{IoU=0.7} \\ 
\cline{3-8} 
 ~ & ~ & Easy & Moderate & Hard  & Easy & Moderate & Hard\\ 
\hline
Mono3D  \cite{chen2016monocular}  
& Mono   & 30.50 & 22.39 & 19.16 &  5.22 &  5.19 &  4.13 \\
Deep3DBox \cite{mousavian20173d} 
& Mono   & 30.02 & 23.77 & 18.83 &  9.99 &  7.71 &  5.30 \\ 
Multi-Fusion \cite{xu2018multi}   
& Mono   & 55.02 & 36.73 & 31.27 & 22.03 & 13.63 & 11.60 \\
ROI-10D~\cite{manhardt2018roi} & Mono  & - & - & -  & 14.76 & 9.55 & 7.57 \\ 
Pseudo-LiDAR~\cite{wang2019pseudo} & Mono  & 70.8 & 49.4 & 42.7  & 40.6 & 26.3 & 22.9 \\ 
Ours  & Mono  & {\bf 72.64} & {\bf 51.82}   & {\bf 44.21} & {\bf 43.75} & {\bf 28.39} & {\bf 23.87} \\  
\hline
\end{tabular}
\end{center}
\caption{3D localization performance: Average Precision ($AP_{loc}$) (in \%) of bird's eye view boxes on KITTI {\it validation} set.}
\label{table:localization}
\end{table*}

\begin{table*}[!ht]
\begin{center}
\begin{tabular}{c||c|c|c|c|c|c|c}
\hline
\multirow{2}{*}{Method} & \multirow{2}{*}{Data} & \multicolumn{3}{c|}{IoU=0.5} & \multicolumn{3}{c}{IoU=0.7} \\ 
\cline{3-8} 
~ & ~ & Easy & Moderate & Hard  & Easy & Moderate & Hard\\ 
\hline
Mono3D \cite{chen2016monocular}  
& Mono    & 25.19 & 18.20 & 15.52 & 2.53 &  2.31 & 2.31 \\
Deep3DBox \cite{mousavian20173d} 
& Mono    & 27.04 & 20.55 & 15.88 & 5.85 &  4.10 & 3.84 \\ 
Multi-Fusion \cite{xu2018multi}   
& Mono    & 47.88 & 29.48 & 26.44 &10.53 & 5.69  & 5.39 \\
ROI-10D~\cite{manhardt2018roi}   & Mono  & - & - & -  & 10.25 & 6.39 & 6.18 \\ 
MonoGRNet~\cite{qin2018monogrnet}   & Mono  & 50.51 & 36.97 & 30.82  & 13.88 & 10.19 & 7.62 \\ 
Pseudo-LiDAR~\cite{wang2019pseudo} & Mono  & 66.3 & 42.3 & 38.5  & 28.2 & 18.5 & 16.4 \\ 
Ours    & Mono       & {\bf 68.86} &  {\bf 49.19}      & {\bf 42.24} & {\bf 32.23} & {\bf 21.09} & {\bf 17.26} \\ 
\hline
\end{tabular}
\end{center}
\caption{3D detection performance: Average Precision ($AP_{3D}$) (in \%) of 3D boxes on KITTI {\it validation} set.}
\label{table:detection}
\end{table*}

\subsection{RGB Information Aggregation}
\begin{figure}[t]
\begin{center}
{\includegraphics[width=0.475\textwidth]{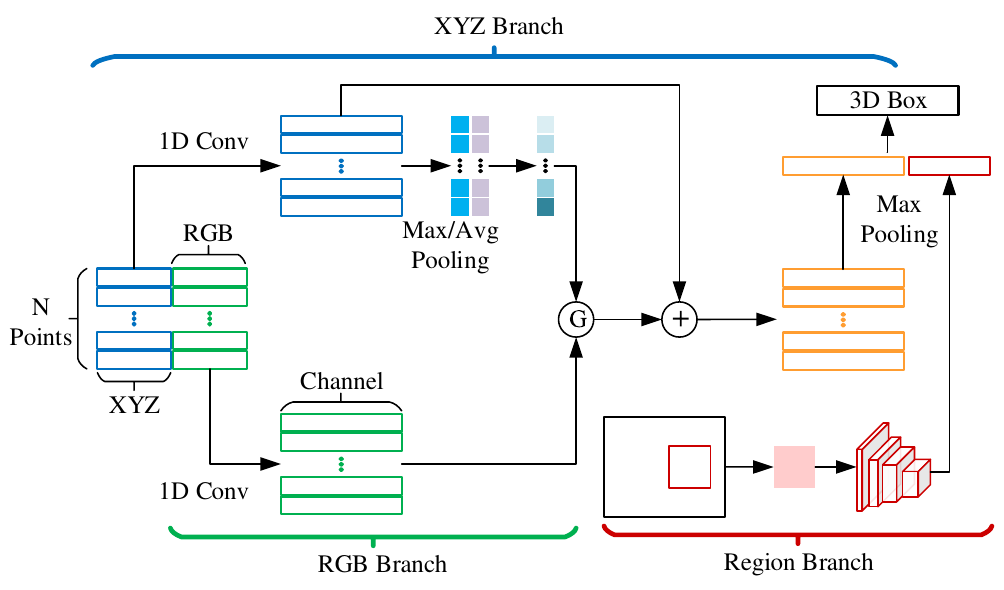}}
\end{center}
\caption{3D box estimation (Det-Net) with RGB features fusion module. {\bf G} is an attention map generated using Eq.~\ref{eq:g}. }
\label{fig:RGB}
\end{figure}
In order to further improve the performance and robustness of our method, we propose to aggregate complementary RGB information to point cloud. 
Specifically, we add RGB information to the generated point cloud by replacing Eq.~\ref{pointset} with:
\begin{equation}
	S = \{\ p  \ | \ p \leftarrow [{\bf F}(v), {\bf D}(v)], \ v\in {\bf B} \} ,
\end{equation}
where {\bf D} is a function which output the corresponding RGB values of input point.
In this way, the points are encoded as 6D vectors: $[x, y, z, r, g, b]$.
However, simply relying on this simple method (we call it 'plain concat' in experiment part) to add RGB information is not feasible.
So, as shown in Fig.~\ref{fig:RGB}, we introduce an attention mechanism for the fusion task.
The attention mechanism has been successfully applied in various tasks such as image caption generation and machine translation for selecting useful information.
Specifically, we utilize the attention mechanism for guiding
the message passing between the spatial features and RGB features. 
Since the passed information flow is not always useful, the attention can act as a gate function to control the flow, in other words to make the network automatically learn to focus or to ignore information from other features.
When we pass RGB message to its corresponding point, an attention map ${\bf G}$ is first produced from the feature maps ${\bf F}$ generated from XYZ branch as follow:
\begin{equation}
	{\bf G} \leftarrow \sigma(f([{\bf F}_{max}^{xyz}, {\bf F}_{avg}^{xyz}])) ,
\label{eq:g}
\end{equation}
where $f$ is the nonlinear function learned from a convolution layer and $\sigma$ is a sigmoid function for normalizing the attention map.
Then the message is passed with the attention map controlled as follow:
\begin{equation}
	{\bf F}^{xyz} \leftarrow {\bf F}^{xyz} + {\bf G}\odot {\bf F}^{rgb} ,
\end{equation}
where $\odot$ denotes element-wise multiplication.
In addition to point-level features fusion, we also introduce another branch to provide object-level RGB information.
In particular, we first crop the RoI from RGB image and resize it to 128$\times$128. 
Then we use a CNN to extract the object-level feature maps ${\bf F}^{obj}$ and the final feature maps set ${\bf F}$ obtained from the fusion module is:
${\bf F} \leftarrow CONCAT({\bf F}^{xyz}, {\bf F}^{obj})$, where $CONCAT$ denotes the concatenation operation.

\subsection{Implementation Details.}
\noindent
\textbf{Optimization.}
The whole training process is performed with two phases.
In the first phase, we only optimize the intermediate nets according to the training strategies of original papers.
After that, we simultaneously optimize the two networks for 3D detection jointly with a multi-task loss function:
\begin{equation}
	L = L_{loc} + L_{det} + \lambda L_{corner} ,
\end{equation}
where $L_{loc}$ is the loss function for the lightweight location net (center only) and $L_{loc}$ is for 3D detection net (center, size and heading angle).
We also use the corner loss \cite{qi2017frustum} where the output targets are first decoded into oriented 3D boxes and then smooth L1 loss is computed on the (x, y, z) coordinates of eight box corners directly with regard to ground truth.
We train the nets for 200 epochs using Adam optimizer with batch size of 32.
The learning rate is initially set to 0.001 and reduced by half for every 20 epochs.
The whole training process can be completed in one day.

\noindent
\textbf{Implementation details.}
The proposed method is implemented base on PyTorch and on Nvidia 1080Ti GPUs.
The two intermediate networks of proposed method naturally supports any network structure. 
We implement some different methods as described in their papers exactly, and the relevant analysis can be found in experimental part.
For the 3D detection nets, we use PointNet as our backbone nets and train them from scratch with random initialization.
Moreover, the dropout strategy with keep rate 0.7 is applied into every fully connected layers except the last one.
For the RGB values, we first normalize the range of them to (0, 1) by dividing 255, and then the data distribution
of each color channel is regularized into standard normal
distribution.
For the region branch in RGB features fusion module, we use ResNet-34 with half channels and global pooling to get the 1$\times$1$\times$256 features.

\section{Experimental Results}
\label{sec:four}

We evaluate our approach on the challenging KITTI dataset \cite{Geiger2012CVPR} which provides 7,481 images for training and 7,518 images for testing.
Detection and localization tasks are evaluated in three regimes: {\it easy}, {\it moderate} and {\it hard}, according to the occlusion and truncation levels of objects. 
Since the ground truth for the test set is not available and the access to the test server is limited, we conduct comprehensive evaluation using the protocol described in  \cite{chen2016monocular, chen20153d, chen2017multi}, and subdivide the training data into a {\it training} set and a {\it validation} set, which results in 3,712 data samples for training and 3,769 data samples for validation. 
The split avoids samples from the same sequence being included in
both {\it training} and {\it validation} set\cite{chen2016monocular}.

\subsection{Comparing with other methods}
\label{sec:4-2}
\noindent
{\bf Baselines.}
As this work aims at monocular 3D object detection, our approach is mainly compared to other methods with only monocular images as input.
Here five methods are chosen for comparisons: Mono3D \cite{chen2016monocular}, Deep3DBox \cite{mousavian20173d} and Multi-Fusion \cite{xu2018multi}, ROI-10D \cite{manhardt2018roi}, MonoGRNet \cite{qin2018monogrnet} and Pseudo-LiDAR~\cite{wang2019pseudo}.

\noindent
{\bf Car.}
The evaluation results of 3D localization and detection tasks on KITTI {\it validation} set are presented in Table \ref{table:localization} and \ref{table:detection}, respectively.
The proposed method consistently outperforms all the competing approaches across all three difficulty levels.
For localization task, the proposed method outperforms Multi-Fusion~\cite{xu2018multi} by $\sim${\bf 15} $AP_{loc}$ in {\it moderate} setting.
For 3D detection task, our method achieves $\sim${\bf 12.2} and $\sim${\bf 10.9} $AP_{3D}$ improvement (moderate) over the recently proposed MonoGRNet~\cite{qin2018monogrnet} under IoU thresholds of 0.5 and 0.7.
In the {\it easy} setting, our improvement is more prominent.
Specifically, our method achieves $\sim${\bf 21.7} and $\sim${\bf 18.4} improvement over previous state-of-the-art on localization and detection tasks (IoU=0.7).
Note that there is no complicated prior knowledge or constraints such as \cite{chen2016monocular,chen20153d,manhardt2018roi}, which strongly confirms the importance of data representation.

Compared with Pseudo-LiDAR~\cite{wang2019pseudo}, which is concurrent to this work, the proposed method has about $\sim${\bf 1.5} $AP$ improvement on each metric. This is because of the modification of the background points segmentation algorithm and the introduction of RGB information. We will discuss this in detail in Sec.~\ref{sec:4-3}.

Table~\ref{table:testing} shows the results on {\it testing} set, and more details, such as Precision-Recall curve, can be found on \href{http://www.cvlibs.net/datasets/kitti/eval_object_detail.php?&result=4de1926d9fa1a857e30cea1c27e6573dc577fc38}{KITTI official server}.
The {\it testing} set results also show the superiority of our method in performance compared with others.

\begin{table}[!ht]
\begin{center}
\begin{tabular}{c|c|c|c|c}
\hline
Method & Task & Easy & Moderate & Hard \\ 
\hline
Multi-Fusion \cite{xu2018multi} & Loc. & 13.73 & 9.62 &  8.22 \\
RoI-10D \cite{manhardt2018roi} & Loc. & 16.77 & 12.40 & 11.39 \\
Ours & Loc. &  {\bf 27.91}  &  {\bf 22.24}  &  {\bf 18.62} \\
\hline
Multi-Fusion \cite{xu2018multi} & Det. &  7.08  &  5.18  &  4.68 \\
RoI-10D \cite{manhardt2018roi} & Det. & 12.30 & 10.30 & 9.39\\
Ours & Det. &  {\bf 21.48}  &  {\bf 16.08}  & {\bf 15.26} \\
\hline
\end{tabular}
\end{center}
\caption{AP(\%) for 3D localization (Loc.) and 3D detection (Det.) tasks on the KITTI {\it testing} set.}
\label{table:testing}
\end{table}

\subsection{Detailed analysis of proposed method}
\label{sec:4-3}
In this section we provide analysis and ablation experiments to validate our design choices.

\begin{figure*}[!t]
\begin{center}
{\includegraphics[width=0.33\textwidth, height=0.105\textwidth]{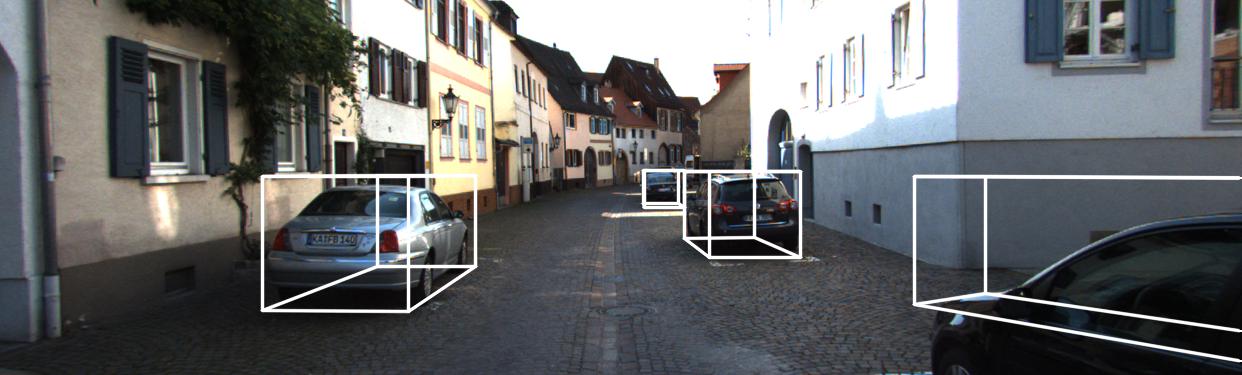}}
\vspace{2pt}
{\includegraphics[width=0.33\textwidth, height=0.105\textwidth]{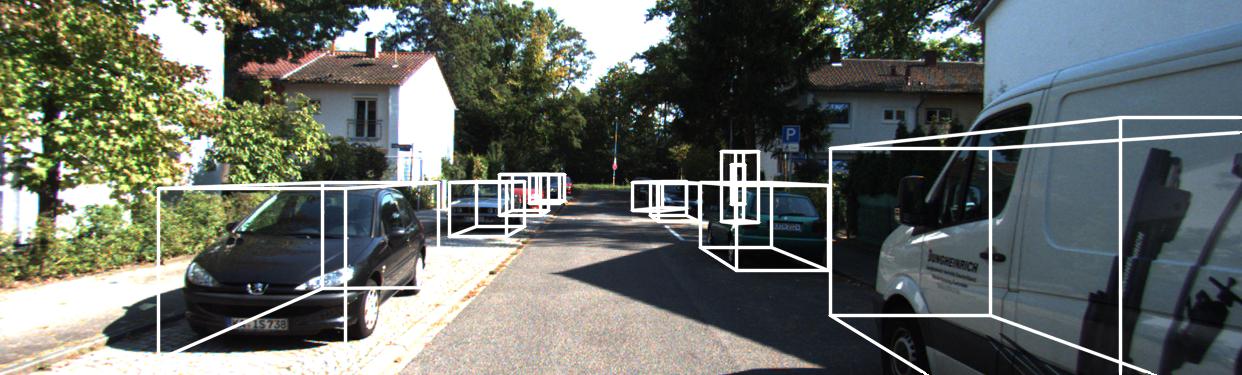}}
{\includegraphics[width=0.33\textwidth, height=0.105\textwidth]{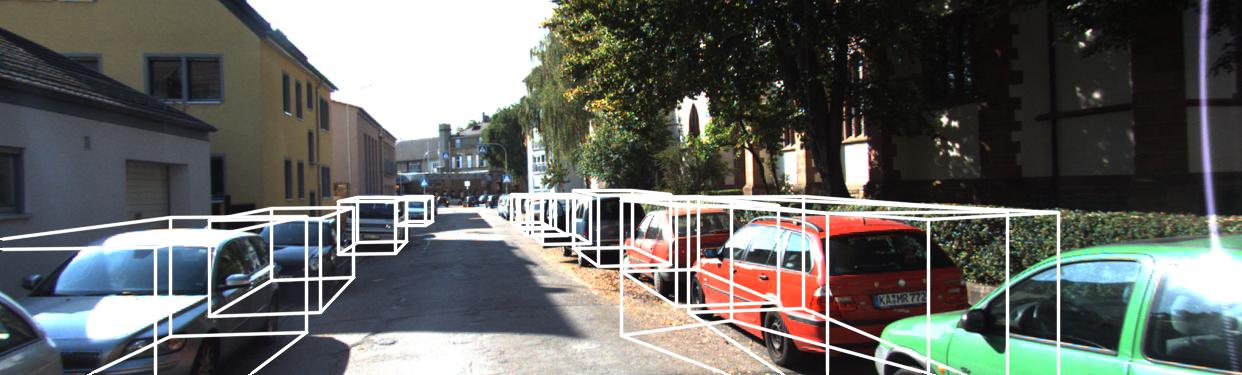}}
\vspace{2pt}
{\includegraphics[width=0.33\textwidth, height=0.105\textwidth]{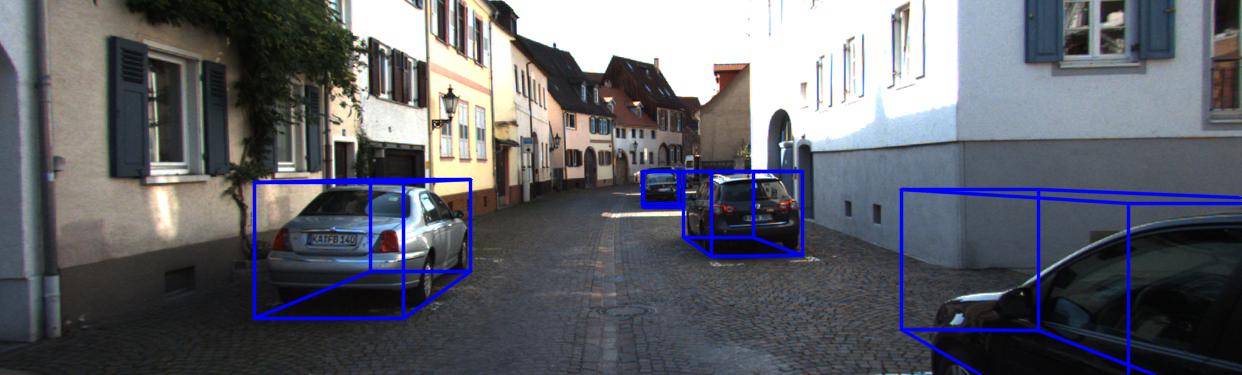}}
{\includegraphics[width=0.33\textwidth, height=0.105\textwidth]{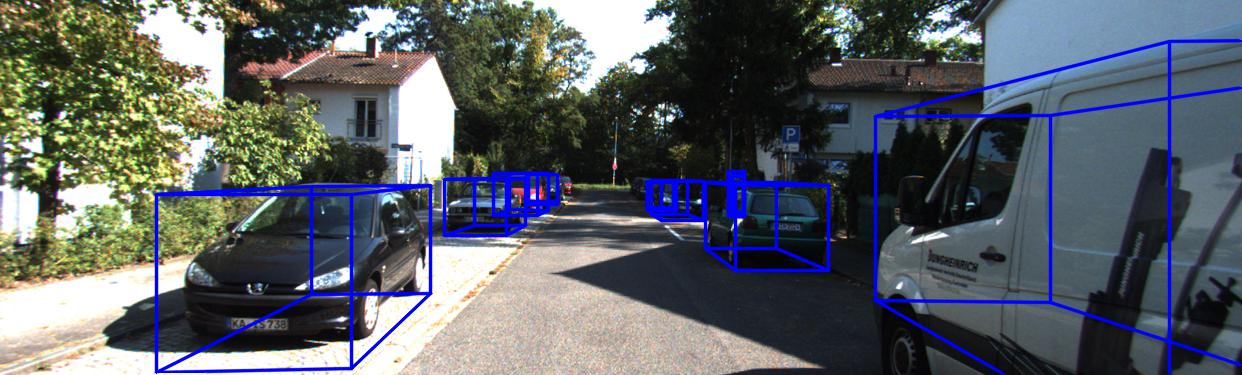}}
{\includegraphics[width=0.33\textwidth, height=0.105\textwidth]{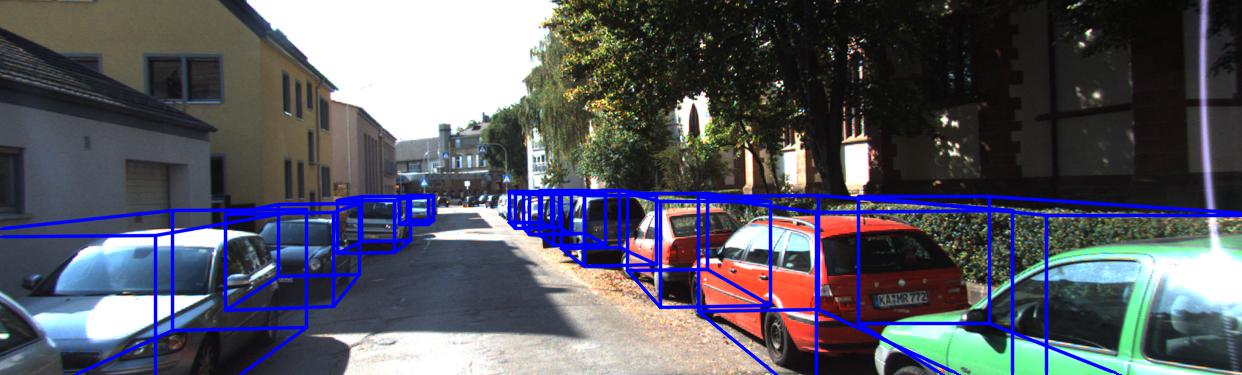}}

\end{center}
\vspace{-5pt}
\caption{{\bf Qualitative comparisons of RGB information:} 3D Boxes are projected to the image plane. 
The detection results using XYZ information only are represented by write boxes, and blue boxes come from the model trained with RGB features fusion module.
The proposed RGB fusion method can improve the 3D detection accuracy, especially for occlusion/truncation cases.
}
\label{fig:rgb}
\vspace{-5pt}
\end{figure*}

\noindent
\textbf{RGB information.}
We further evaluate the effect of the proposed RGB fusion module, and the baselines are the proposed method without RGB values and using them as additional channels of generated points.
Table~\ref{table:rgb} shows the relevant results for {\it Car} category on KITTI.
It can be seen that the proposed module obtains around {\bf 2.1} and {\bf 1.6} mAP improvement (moderate) on localization and detection task, respectively. The qualitative comparisons can be found in Fig~\ref{fig:rgb}.
Quantitative and qualitative results both show the effectiveness of proposed RGB fusion module.
Besides, one thing to note is that incorrect use of RGB information such as plain concat will lead to performance degradation.

\begin{table}[!ht]
\begin{center}
\begin{tabular}{c||c|c|c|c}
\hline
 ~  & Task & Easy & Moderate & Hard \\ 
\hline
w/o RGB & Loc. & 41.29 & 26.28 & 22.75 \\
plain concat   & Loc. & 36.17 & 25.34 & 21.94 \\
ours    & Loc. & {\bf 43.75} & {\bf 28.39} & {\bf 23.87} \\
\hline
w/o RGB & Det. & 30.73 & 19.46 & 16.72 \\
plain concat   & Det. &  27.20 & 18.25 & 16.15 \\
ours    & Det. & {\bf 32.23} & {\bf 21.09} & {\bf 17.26} \\
\hline
\end{tabular}
\end{center}
\caption{Ablation study of RGB information. The metric is $AP_{3D}^{0.7}$ on KITTI {\it validation} set.}
\label{table:rgb}
\end{table}

\noindent
\textbf{Points segmentation.}
We compare the proposed points segmentation method and the 3D segmentation PointNet which is used in \cite{qi2017frustum}. 
The baseline is to estimate 3D boxes directly using point clouds with noise which can be regarded as all points are classified into positive samples.
As shown in Table \ref{table:segmentation}, our prior-based method outperforms baseline and segmentation PointNet obviously which proves the effectiveness of the proposed method and 
Table~\ref{table:threshold} shows that the proposed method is robust for varying thresholds.
Meanwhile, the experimental results also show that the learning-based method is not applicable to approximate point clouds segmentation task because it's difficult to obtain reliable labels.
Besides, the proposed method is also much faster than segmentation PointNet (about 5ms on CPU {\it v.s.} 20ms on GPU for each proposal).

\begin{table}[!ht]
\begin{center}
\begin{tabular}{c||c|c|c|c}
\hline
~ & IoU & Easy & Moderate & Hard  \\ 
\hline
w/o segmentation & 0.5 & 66.42 & 44.53 & 40.60  \\
seg-net used in \cite{qi2017frustum} & 0.5 & 67.01 & 45.51 & 40.65   \\ 
ours & 0.5 & {\bf 68.86} & {\bf 49.19} & {\bf 42.24} \\ 
\hline
w/o segmentation  & 0.7 & 27.04 & 18.22 & 16.13 \\
seg-net used in \cite{qi2017frustum} & 0.7  & 29.49 & 18.70 & 16.57\\ 
ours & 0.7 & {\bf 32.23} & {\bf 21.09} & {\bf 17.26} \\ 
\hline
\end{tabular}
\end{center}
\caption{Ablation study of points segmentation. The metric is $AP_{3D}^{0.7}$ on KITTI {\it validation} set.}
\label{table:segmentation}
\end{table}

\begin{table}[!ht]
\begin{center}
\begin{tabular}{c||c|c|c}
\hline
 r & Easy & Moderate & Hard \\ 
\hline
-0.5 & 31.13 & 20.01 & 16.81 \\
0.0  & 31.87 & 20.55 & 17.03 \\
0.5  & {\bf 32.23} & {\bf 21.09} & {\bf 17.26} \\
1.0  & 31.93 & 20.93 & 17.18 \\
\hline
\end{tabular}
\end{center}
\caption{$AP_{3D}^{0.7}$(\%) of different points segmentation threshold $r$ (in meters) for 3D detection on the KITTI {\it validation} set.}
\label{table:threshold}
\end{table}

\noindent
\textbf{Depth maps.}
As described in Sec. \ref{sec:three}, our approach depends on the point clouds generated from the output of depth generator.
In order to study the impact of quality of depth maps on the overall performance of proposed method, we implemented four different depth generators \cite{godard2017unsupervised, liang2018deep, mayer2016large, chang2018pyramid}.  
From the results shown in Table \ref{table:depth}, we find that 3D detection accuracy increases significantly when using more accurate depth (more details about the accuracy of depth maps can be found in the supplement material).
It's worth noting that even if we use the unsupervised monocular depth generator \cite{godard2017unsupervised}, the proposed method still outperforms  \cite{manhardt2018roi} by a large margin.

\begin{table}[!ht]
\begin{center}
\begin{tabular}{c||c|c|c|c}
\hline
Depth & Task & Easy & Mod. & Hard \\ 
\hline
MonoDepth\cite{godard2017unsupervised} &
Loc. & 32.42 & 20.26 & 17.21 \\
DORN\cite{liang2018deep} &
Loc. & 43.75 & 28.39 & 23.87 \\
DispNet\cite{mayer2016large} &
Loc. & 47.41 & 30.72 & 25.66 \\
PSMNet \cite{chang2018pyramid} &
Loc. & 60.18 & 34.01 & 30.32 \\
\hline
MonoDepth\cite{godard2017unsupervised} &
Det. & 23.12 & 15.45 & 14.19 \\
DORN\cite{liang2018deep} &
Det. & 32.23 & 21.09 & 17.26 \\
DispNet\cite{mayer2016large} &
Det. & 36.97 & 23.69 & 19.25 \\
PSMNet \cite{chang2018pyramid} &
Det. & 45.85 & 26.03 & 23.16 \\
\hline
\end{tabular}
\end{center}
\caption{Comparisons of different depth generators.
Metrics are $AP_{loc}^{0.7}$ and $AP_{3D}^{0.7}$ on KITTI {\it validation} set.}
\label{table:depth}
\end{table}

\noindent
\textbf{Sampling quantity.}
Some studies such as \cite{qi2017pointnet, qi2017pointnet++} observe that classification/segmentation accuracy will decrease dramatically as the number of points decreases, and we will show that our approach is not so sensitive to the number of points.
In our approach, we randomly select a fixed number (512 points for default configuration) of point clouds to do 3D detection task.
Table. \ref{table:sampling} shows the performance of proposed method under different sampling quantity.
According to the results, $AP_{3D}$ will increase as the number of points increases at the beginning.
Then, after reaching a certain level ($\sim$512 points), the performance tends to be stable.
It is worth noting that we still get a relatively good detection performance even if there are few sampling points.

\begin{table}[!ht]
\begin{center}
\begin{tabular}{c||c|c|c}
\hline
Sampling Quantity & Easy & Mod. & Hard\\ 
\hline
64     & 27.91 & 19.41 & 16.31  \\
128    & 29.72 & 19.62 & 16.64  \\
256    & 30.99 & 20.71 & 17.18  \\
512    & {\bf 32.23} & {\bf 21.09} & {\bf 17.26}  \\
1024   & 31.44 & 21.01 & 17.23  \\
\hline
\end{tabular}
\end{center}
\caption{Comparisons of  different sampling quantity.
The metric is $AP_{3D}^{0.7}$(\%) on KITTI {\it validation} set.
Note that the number of sample points is consistent at the training and testing phase.}
\label{table:sampling}
\end{table}

\begin{figure}[h]
\begin{center}
{\includegraphics[width=0.23\textwidth]{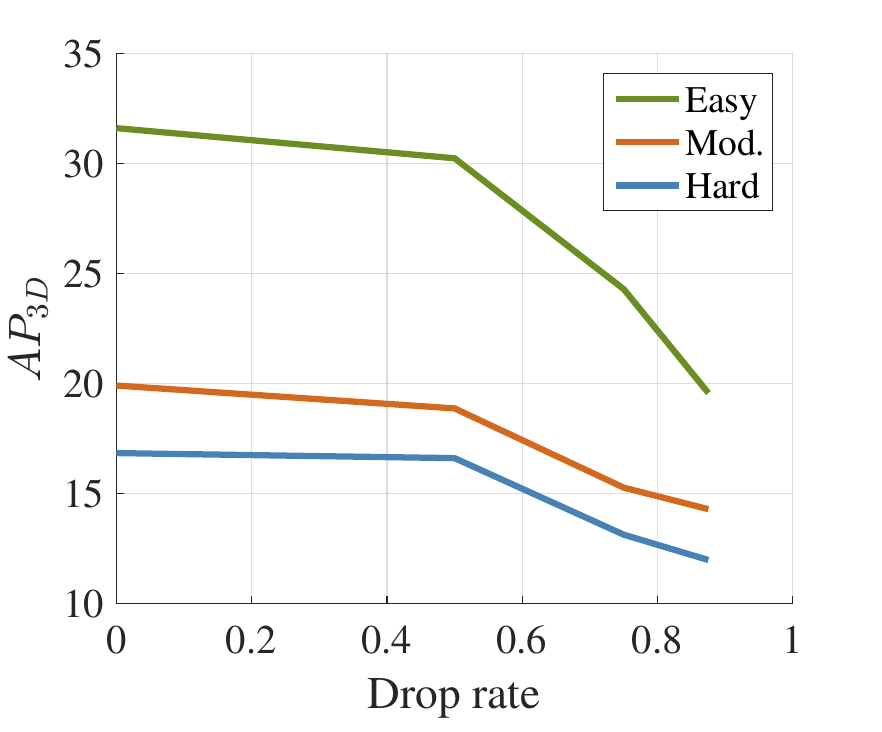}
\label{fig:drop}}
{\includegraphics[width=0.23\textwidth]{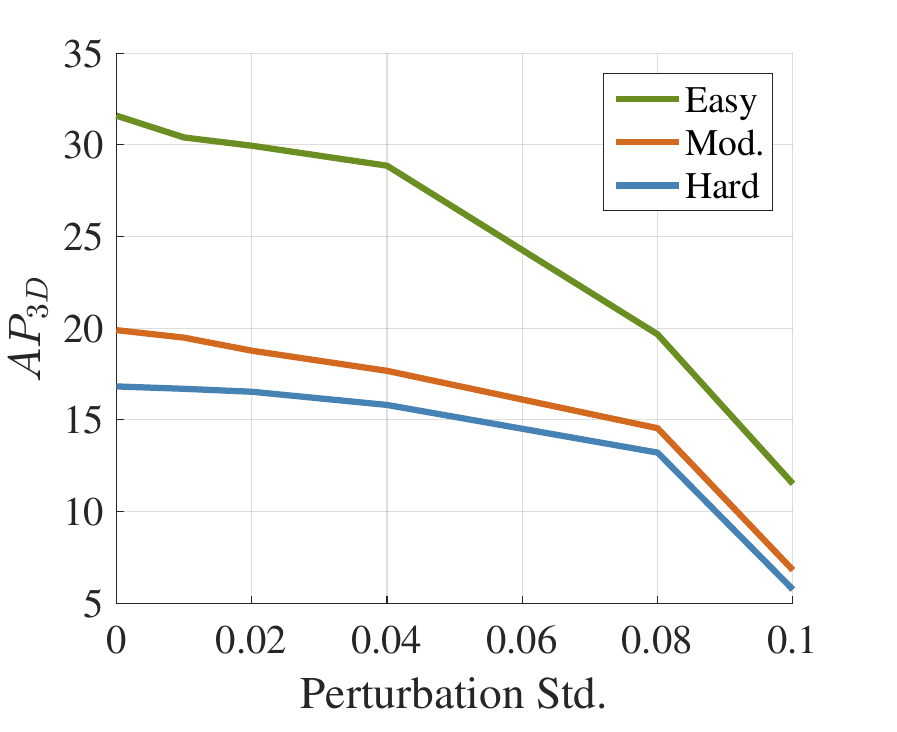}
\label{fig:perturbation}}
\end{center}
\caption{{\it Left}: robustness test of random point dropout.
{\it Right}: robustness test of random perturbations (Gaussian noise is added into each point independently).
The metric is $AP_{3D}^{0.7}$(\%) for {\it Car} on KITTI {\it validation} set.
}
\vspace{-5pt}
\label{fig:robustness}
\end{figure}

\begin{figure*}[!t]
\begin{center}
{\includegraphics[width=0.33\textwidth, height=0.105\textwidth]{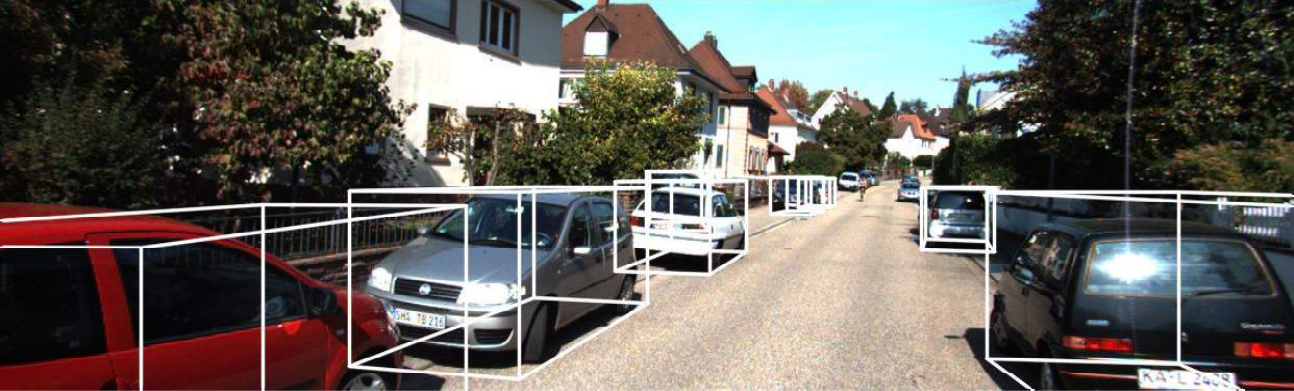}}
\vspace{2pt}
{\includegraphics[width=0.33\textwidth, height=0.105\textwidth]{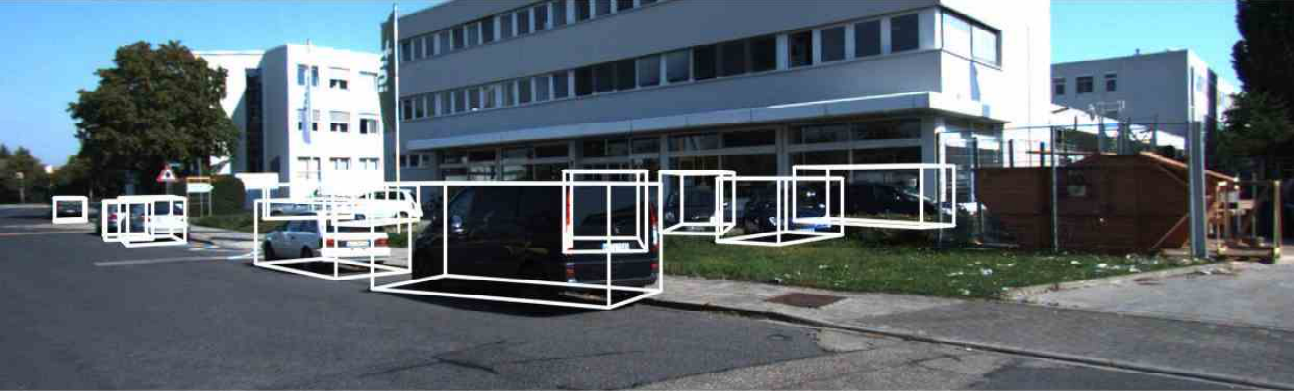}}
{\includegraphics[width=0.33\textwidth, height=0.105\textwidth]{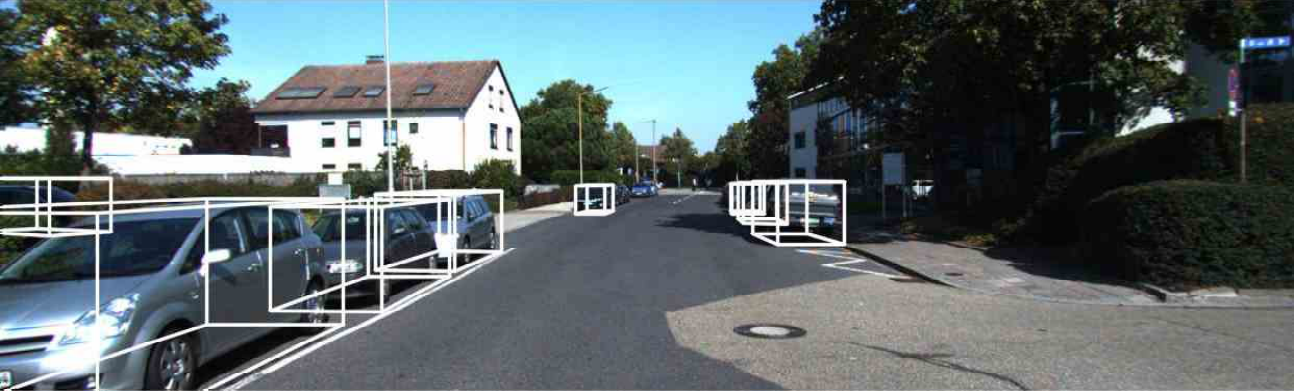}}
\vspace{2pt}
{\includegraphics[width=0.33\textwidth, height=0.105\textwidth]{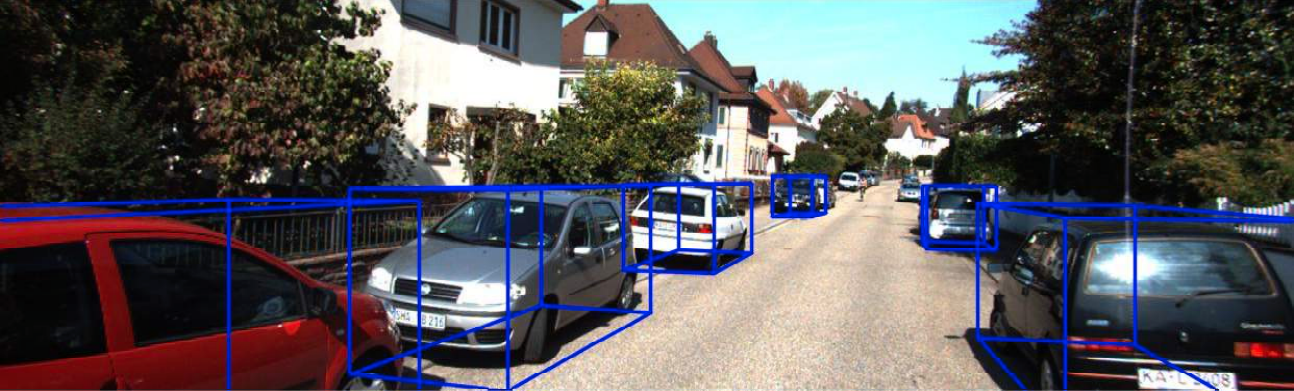}}
{\includegraphics[width=0.33\textwidth, height=0.105\textwidth]{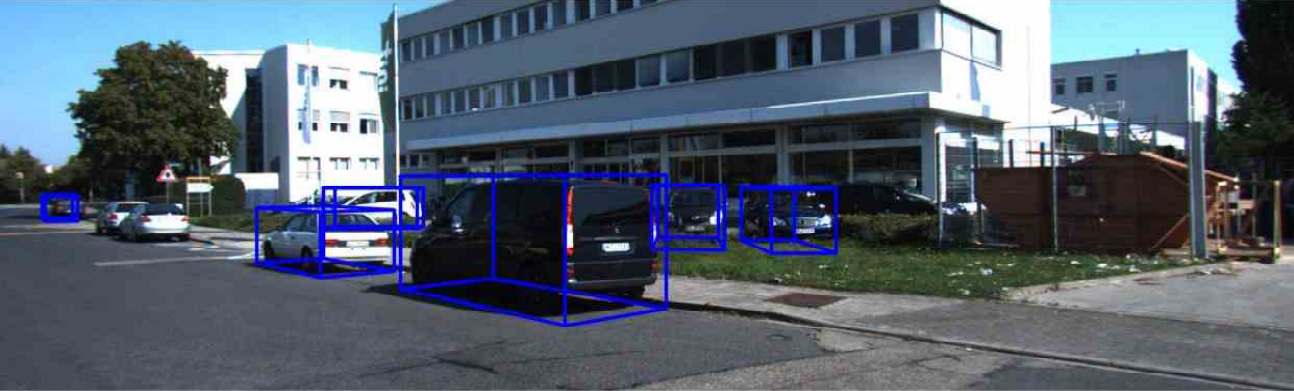}}
{\includegraphics[width=0.33\textwidth, height=0.105\textwidth]{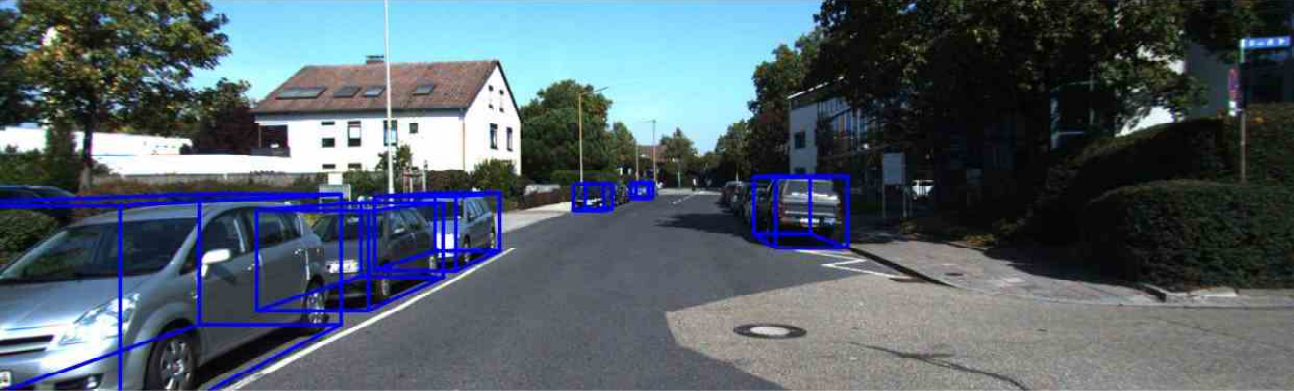}}
\vspace{2pt}
{\includegraphics[width=0.33\textwidth, height=0.105\textwidth]{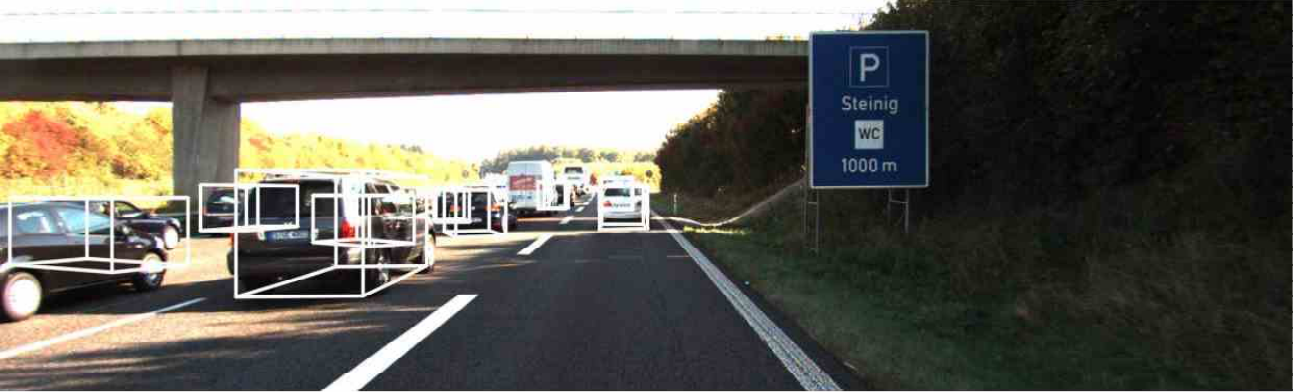}}
{\includegraphics[width=0.33\textwidth, height=0.105\textwidth]{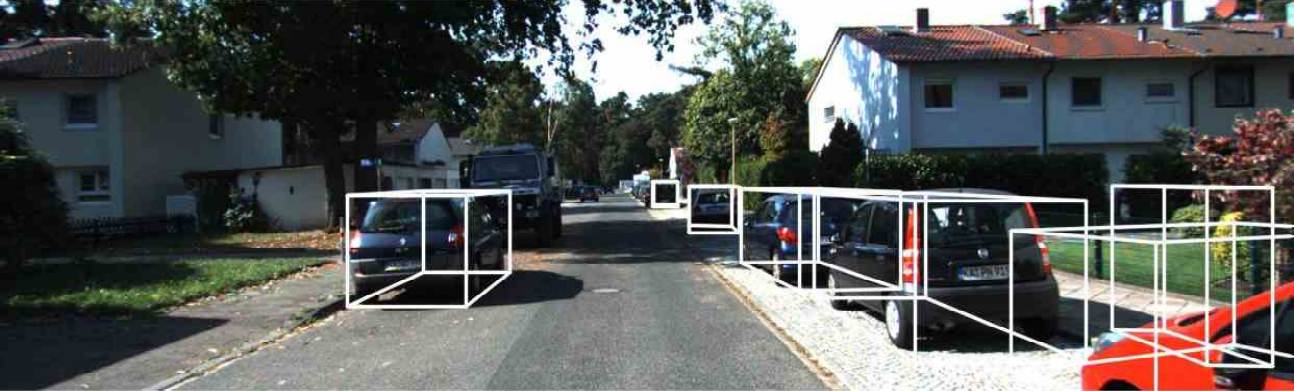}}
{\includegraphics[width=0.33\textwidth, height=0.105\textwidth]{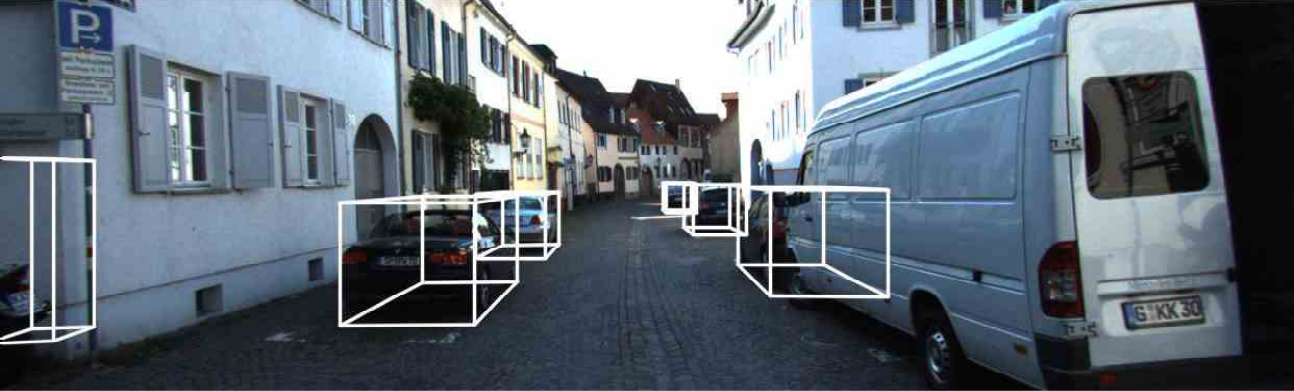}}
{\includegraphics[width=0.33\textwidth, height=0.105\textwidth]{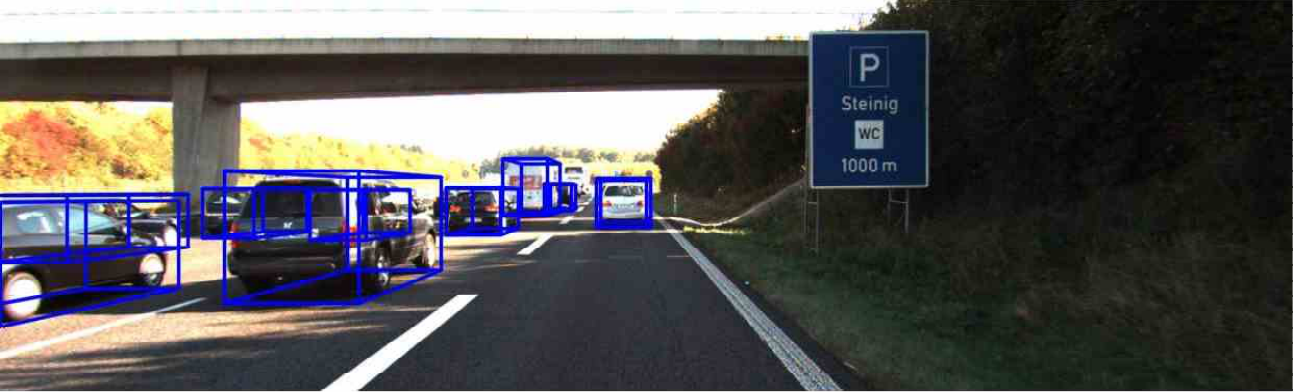}}
{\includegraphics[width=0.33\textwidth, height=0.105\textwidth]{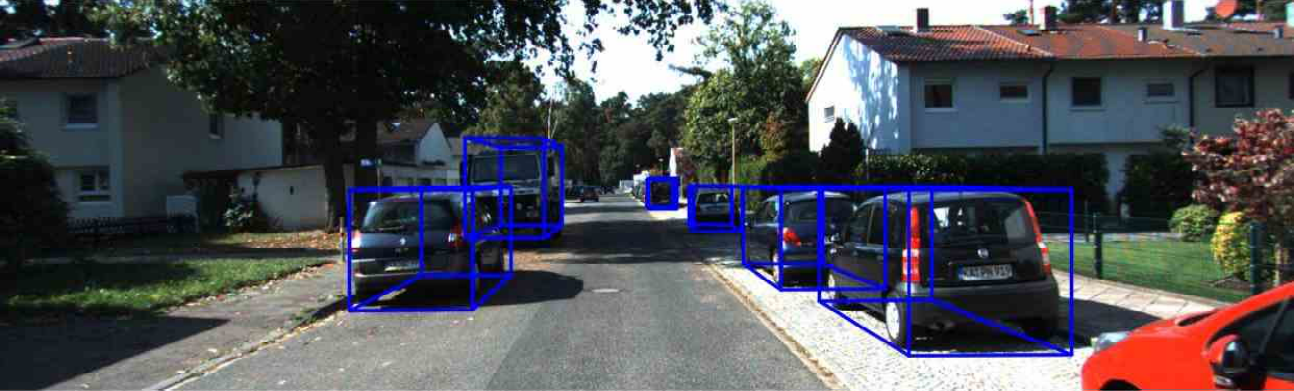}}
{\includegraphics[width=0.33\textwidth, height=0.105\textwidth]{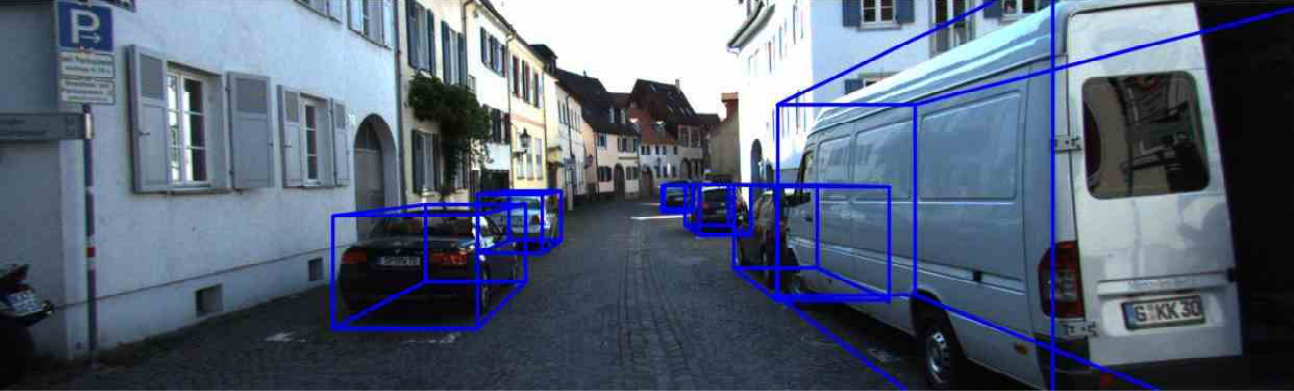}}
\end{center}
\vspace{-5pt}
\caption{{\bf Qualitative comparisons of 3D detection results:} 3D Boxes are projected to the image plane. White boxes represent our predictions, and  blue boxes come from ground truth.
}
\label{fig:qualitative}
\end{figure*}

\begin{figure}[t]
\begin{center}
{\includegraphics[width=0.475\textwidth]{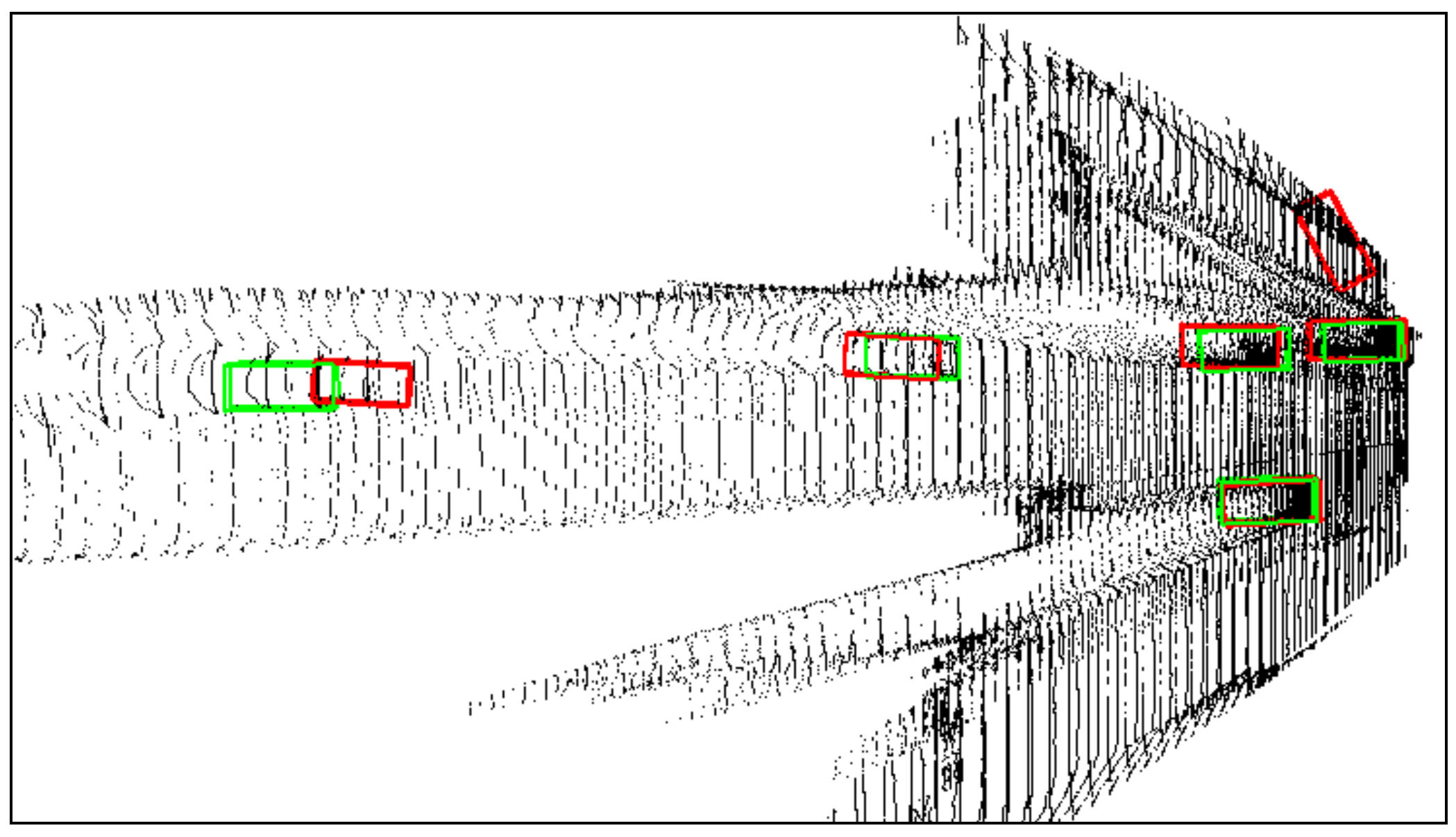}}
\end{center}
\vspace{-5pt}
\caption{{\bf A qualitative result of 3D localization :} 3D Boxes are projected to the ground plane. Red boxes represent our predictions, and green boxes come from ground truth.
}
\label{fig:localization}
\end{figure}

\noindent
{\bf Robustness.}
We show that the proposed method is robust to various kinds of input corruptions.
We first set the sampling quantity to 512 in training phase, but use different values in the testing phase.
Fig. \ref{fig:drop} shows that the proposed method has more than 70\% $AP_{3D}$ even when 80\% of the points are missed.
Then, we test the robustness of model to point perturbations, and the results are shown in Fig \ref{fig:perturbation}.

\noindent
\textbf{Network architecture.}
We also investigate the impact of different 3D detection network architectures on overall performance (the previously reported results are all based on PointNet), and the experimental result are shown in Table.~\ref{table:nets}.

\begin{table}[!ht]
\begin{center}
\begin{tabular}{c||c|c|c|c}
\hline
 ~ & Data & Easy & Mod. & Hard  \\ 
\hline
PointNet~\cite{qi2017pointnet} & Mono & 32.23 & 21.09 & 17.26  \\
PointNet++~\cite{qi2017pointnet++} & Mono & 33.17 & 21.71 & 17.61\\
RSNet \cite{huang2018recurrent} & Mono & 33.93 & 22.34 & 17.79  \\
\hline
\end{tabular}
\end{center}
\caption{Comparisons of different 3D detection network architectures.
The metric is $AP_{3D}^{0.7}$(\%) on KITTI {\it validation} set.}
\vspace{-5pt}
\label{table:nets}
\end{table}

\subsection{Qualitative Results and Failure Mode}

We visualize some detection results of our approach in Fig. \ref{fig:qualitative} and a typical localization result in Fig. \ref{fig:localization}.
In general, our algorithm can get a good detection result. However, because it's a 2D-driven framework, the proposed method will fail if the 2D box is a false positive sample or missing. Besides, for distant objects, our algorithm is difficult to give accurate results because the depth is not reliable (the leftmost car in Fig. \ref{fig:localization} is 70.35 meters away from the camera).

\section{Conclusions}
\label{sec:five}

We proposed a framework for accurate 3D object detection with monocular images in this paper.
Unlike other image-based methods, our method solves this problem in the reconstructed 3D space in order to exploit 3D contexts explicitly.
We argue that the point cloud representation is more suitable for 3D related tasks than depth maps.
Besides, we propose a multi-modal feature fusion module to embed the complementary RGB cue into the generated point clouds representation to enhance the discriminative capability of generated point clouds.
Our approach significantly outperforms existing monocular-based method for 3D localization and detection tasks on KITTI benchmark.
In addition, the extended versions verifies the design strategy can also be applied to stereo-based and LiDAR-based methods.

\section*{Acknowledgments}
This work was supported in part by the National Natural Science Foundation of China (NSFC) under Grants No. 61772108, 61572096 and No. 61733002.

{\small
\bibliographystyle{ieee_fullname}
\bibliography{egbib}

\begin{thebibliography}{10}\itemsep=-1pt

\bibitem{chabot2017deep}
Florian Chabot, Mohamed Chaouch, Jaonary Rabarisoa, C{\'e}line Teuli{\`e}re,
  and Thierry Chateau.
\newblock Deep manta: A coarse-to-fine many-task network for joint 2d and 3d
  vehicle analysis from monocular image.
\newblock In {\em The IEEE Conference on Computer Vision and Pattern
  Recognition (CVPR)}, pages 2040--2049, 2017.

\bibitem{chang2018pyramid}
Jia-Ren Chang and Yong-Sheng Chen.
\newblock Pyramid stereo matching network.
\newblock In {\em The IEEE Conference on Computer Vision and Pattern
  Recognition (CVPR)}, pages 5410--5418, 2018.

\bibitem{chen2016monocular}
Xiaozhi Chen, Kaustav Kundu, Ziyu Zhang, Huimin Ma, Sanja Fidler, and Raquel
  Urtasun.
\newblock Monocular 3d object detection for autonomous driving.
\newblock In {\em The IEEE Conference on Computer Vision and Pattern
  Recognition (CVPR)}, pages 2147--2156, 2016.

\bibitem{chen20153d}
Xiaozhi Chen, Kaustav Kundu, Yukun Zhu, Andrew~G Berneshawi, Huimin Ma, Sanja
  Fidler, and Raquel Urtasun.
\newblock 3d object proposals for accurate object class detection.
\newblock In {\em Advances in Neural Information Processing Systems}, pages
  424--432, 2015.

\bibitem{chen2017multi}
Xiaozhi Chen, Huimin Ma, Ji Wan, Bo Li, and Tian Xia.
\newblock Multi-view 3d object detection network for autonomous driving.
\newblock In {\em The IEEE Conference on Computer Vision and Pattern
  Recognition (CVPR)}, volume~1, page~3, 2017.

\bibitem{fu2018deep}
Huan Fu, Mingming Gong, Chaohui Wang, Kayhan Batmanghelich, and Dacheng Tao.
\newblock Deep ordinal regression network for monocular depth estimation.
\newblock In {\em The IEEE Conference on Computer Vision and Pattern
  Recognition (CVPR)}, pages 2002--2011, 2018.

\bibitem{Geiger2012CVPR}
Andreas Geiger, Philip Lenz, and Raquel Urtasun.
\newblock Are we ready for autonomous driving? the kitti vision benchmark
  suite.
\newblock In {\em The IEEE Conference on Computer Vision and Pattern
  Recognition (CVPR)}, 2012.

\bibitem{girshick2015fast}
Ross Girshick.
\newblock Fast r-cnn.
\newblock In {\em The IEEE International Conference on Computer Vision (ICCV)},
  pages 1440--1448, 2015.

\bibitem{girshick2014rich}
Ross Girshick, Jeff Donahue, Trevor Darrell, and Jitendra Malik.
\newblock Rich feature hierarchies for accurate object detection and semantic
  segmentation.
\newblock In {\em The IEEE Conference on Computer Vision and Pattern
  Recognition (CVPR)}, pages 580--587, 2014.

\bibitem{godard2017unsupervised}
Cl{\'e}ment Godard, Oisin Mac~Aodha, and Gabriel~J Brostow.
\newblock Unsupervised monocular depth estimation with left-right consistency.
\newblock In {\em The IEEE Conference on Computer Vision and Pattern
  Recognition (CVPR)}, volume~2, page~7, 2017.

\bibitem{he2017mask}
Kaiming He, Georgia Gkioxari, Piotr Doll{\'a}r, and Ross Girshick.
\newblock Mask r-cnn.
\newblock In {\em The IEEE International Conference on Computer Vision (ICCV)},
  pages 2980--2988. IEEE, 2017.

\bibitem{he2016deep}
Kaiming He, Xiangyu Zhang, Shaoqing Ren, and Jian Sun.
\newblock Deep residual learning for image recognition.
\newblock In {\em The IEEE Conference on Computer Vision and Pattern
  Recognition (CVPR)}, pages 770--778, 2016.

\bibitem{huang2018recurrent}
Qiangui Huang, Weiyue Wang, and Ulrich Neumann.
\newblock Recurrent slice networks for 3d segmentation of point clouds.
\newblock In {\em The IEEE Conference on Computer Vision and Pattern
  Recognition (CVPR)}, pages 2626--2635, 2018.

\bibitem{li20173d}
Bo Li.
\newblock 3d fully convolutional network for vehicle detection in point cloud.
\newblock In {\em IEEE International Conference on Intelligent Robots and
  Systems (IROS)}, pages 1513--1518. IEEE, 2017.

\bibitem{Li_2019_CVPR}
Buyu Li, Wanli Ouyang, Lu Sheng, Xingyu Zeng, and Xiaogang Wang.
\newblock Gs3d: An efficient 3d object detection framework for autonomous
  driving.
\newblock In {\em The IEEE Conference on Computer Vision and Pattern
  Recognition (CVPR)}, June 2019.

\bibitem{liang2018deep}
Ming Liang, Bin Yang, Shenlong Wang, and Raquel Urtasun.
\newblock Deep continuous fusion for multi-sensor 3d object detection.
\newblock In {\em Proceedings of the European Conference on Computer Vision
  (ECCV)}, pages 641--656, 2018.

\bibitem{lin2018focal}
Tsung-Yi Lin, Priyal Goyal, Ross Girshick, Kaiming He, and Piotr Doll{\'a}r.
\newblock Focal loss for dense object detection.
\newblock {\em IEEE transactions on pattern analysis and machine intelligence},
  2018.

\bibitem{liu2018deep}
Li Liu, Wanli Ouyang, Xiaogang Wang, Paul Fieguth, Jie Chen, Xinwang Liu, and
  Matti Pietik{\"a}inen.
\newblock Deep learning for generic object detection: A survey.
\newblock {\em arXiv preprint arXiv:1809.02165}, 2018.

\bibitem{luo2018fast}
Wenjie Luo, Bin Yang, and Raquel Urtasun.
\newblock Fast and furious: Real time end-to-end 3d detection, tracking and
  motion forecasting with a single convolutional net.
\newblock In {\em The IEEE Conference on Computer Vision and Pattern
  Recognition (CVPR)}, pages 3569--3577, 2018.

\bibitem{manhardt2018roi}
Fabian Manhardt, Wadim Kehl, and Adrien Gaidon.
\newblock Roi-10d: Monocular lifting of 2d detection to 6d pose and metric
  shape.
\newblock In {\em The IEEE Conference on Computer Vision and Pattern
  Recognition (CVPR)}, June 2019.

\bibitem{maturana2015voxnet}
Daniel Maturana and Sebastian Scherer.
\newblock Voxnet: A 3d convolutional neural network for real-time object
  recognition.
\newblock In {\em IEEE International Conference on Intelligent Robots and
  Systems (IROS)}, pages 922--928. IEEE, 2015.

\bibitem{mayer2016large}
Nikolaus Mayer, Eddy Ilg, Philip Hausser, Philipp Fischer, Daniel Cremers,
  Alexey Dosovitskiy, and Thomas Brox.
\newblock A large dataset to train convolutional networks for disparity,
  optical flow, and scene flow estimation.
\newblock In {\em The IEEE Conference on Computer Vision and Pattern
  Recognition (CVPR)}, pages 4040--4048, 2016.

\bibitem{mousavian20173d}
Arsalan Mousavian, Dragomir Anguelov, John Flynn, and Jana Ko{\v{s}}eck{\'a}.
\newblock 3d bounding box estimation using deep learning and geometry.
\newblock In {\em The IEEE Conference on Computer Vision and Pattern
  Recognition (CVPR)}, pages 5632--5640. IEEE, 2017.

\bibitem{Ouyang_2017_ICCV}
Wanli Ouyang, Kun Wang, Xin Zhu, and Xiaogang Wang.
\newblock Chained cascade network for object detection.
\newblock In {\em The IEEE International Conference on Computer Vision (ICCV)},
  Oct 2017.

\bibitem{ouyang2013joint}
Wanli Ouyang and Xiaogang Wang.
\newblock Joint deep learning for pedestrian detection.
\newblock In {\em Proceedings of the IEEE International Conference on Computer
  Vision}, pages 2056--2063, 2013.

\bibitem{ouyang2017jointly}
Wanli Ouyang, Hui Zhou, Hongsheng Li, Quanquan Li, Junjie Yan, and Xiaogang
  Wang.
\newblock Jointly learning deep features, deformable parts, occlusion and
  classification for pedestrian detection.
\newblock {\em IEEE transactions on pattern analysis and machine intelligence},
  40(8):1874--1887, 2017.

\bibitem{qi2017frustum}
Charles~R. Qi, Wei Liu, Chenxia Wu, Hao Su, and Leonidas~J. Guibas.
\newblock Frustum pointnets for 3d object detection from rgb-d data.
\newblock In {\em The IEEE Conference on Computer Vision and Pattern
  Recognition (CVPR)}, June 2018.

\bibitem{qi2017pointnet}
Charles~R. Qi, Hao Su, Kaichun Mo, and Leonidas~J. Guibas.
\newblock Pointnet: Deep learning on point sets for 3d classification and
  segmentation.
\newblock In {\em The IEEE Conference on Computer Vision and Pattern
  Recognition (CVPR)}, July 2017.

\bibitem{qi2017pointnet++}
Charles~Ruizhongtai Qi, Li Yi, Hao Su, and Leonidas~J Guibas.
\newblock Pointnet++: Deep hierarchical feature learning on point sets in a
  metric space.
\newblock In {\em Advances in Neural Information Processing Systems}, pages
  5099--5108, 2017.

\bibitem{qin2018monogrnet}
Zengyi Qin, Jinglu Wang, and Yan Lu.
\newblock Monogrnet: A geometric reasoning network for monocular 3d object
  localization.
\newblock In {\em Proceedings of the AAAI Conference on Artificial
  Intelligence}, volume~33, pages 8851--8858, 2019.

\bibitem{Ren_2017_CVPR}
Jimmy Ren, Xiaohao Chen, Jianbo Liu, Wenxiu Sun, Jiahao Pang, Qiong Yan,
  Yu-Wing Tai, and Li Xu.
\newblock Accurate single stage detector using recurrent rolling convolution.
\newblock In {\em The IEEE Conference on Computer Vision and Pattern
  Recognition (CVPR)}, July 2017.

\bibitem{ren2015faster}
Shaoqing Ren, Kaiming He, Ross Girshick, and Jian Sun.
\newblock Faster r-cnn: Towards real-time object detection with region proposal
  networks.
\newblock In {\em Advances in neural information processing systems}, pages
  91--99, 2015.

\bibitem{ren2016three}
Zhile Ren and Erik~B Sudderth.
\newblock Three-dimensional object detection and layout prediction using clouds
  of oriented gradients.
\newblock In {\em The IEEE Conference on Computer Vision and Pattern
  Recognition (CVPR)}, pages 1525--1533, 2016.

\bibitem{simonyan2014very}
Karen Simonyan and Andrew Zisserman.
\newblock Very deep convolutional networks for large-scale image recognition.
\newblock {\em arXiv preprint arXiv:1409.1556}, 2014.

\bibitem{song2014sliding}
Shuran Song and Jianxiong Xiao.
\newblock Sliding shapes for 3d object detection in depth images.
\newblock In {\em European conference on computer vision}, pages 634--651.
  Springer, 2014.

\bibitem{sun2018fishnet}
Shuyang Sun, Jiangmiao Pang, Jianping Shi, Shuai Yi, and Wanli Ouyang.
\newblock Fishnet: A versatile backbone for image, region, and pixel level
  prediction.
\newblock In {\em Advances in Neural Information Processing Systems}, pages
  754--764, 2018.

\bibitem{wang2019pseudo}
Yan Wang, Wei-Lun Chao, Divyansh Garg, Bharath Hariharan, Mark Campbell, and
  Kilian~Q Weinberger.
\newblock Pseudo-lidar from visual depth estimation: Bridging the gap in 3d
  object detection for autonomous driving.
\newblock In {\em Proceedings of the IEEE Conference on Computer Vision and
  Pattern Recognition}, pages 8445--8453, 2019.

\bibitem{xu2018multi}
Bin Xu and Zhenzhong Chen.
\newblock Multi-level fusion based 3d object detection from monocular images.
\newblock In {\em The IEEE Conference on Computer Vision and Pattern
  Recognition (CVPR)}, June 2018.

\bibitem{xu2018pad}
Dan Xu, Wanli Ouyang, Xiaogang Wang, and Nicu Sebe.
\newblock Pad-net: Multi-tasks guided prediction-and-distillation network for
  simultaneous depth estimation and scene parsing.
\newblock 2018.

\bibitem{xu2018monocular}
Dan Xu, Elisa Ricci, Wanli Ouyang, Xiaogang Wang, and Nicu Sebe.
\newblock Monocular depth estimation using multi-scale continuous crfs as
  sequential deep networks.
\newblock {\em IEEE Transactions on Pattern Analysis and Machine Intelligence},
  2018.

\bibitem{yang2018pixor}
Bin Yang, Wenjie Luo, and Raquel Urtasun.
\newblock Pixor: Real-time 3d object detection from point clouds.
\newblock In {\em The IEEE Conference on Computer Vision and Pattern
  Recognition (CVPR)}, June 2018.

\bibitem{zeng2017crafting}
Xingyu Zeng, Wanli Ouyang, Junjie Yan, Hongsheng Li, Tong Xiao, Kun Wang, Yu
  Liu, Yucong Zhou, Bin Yang, Zhe Wang, et~al.
\newblock Crafting gbd-net for object detection.
\newblock {\em IEEE transactions on pattern analysis and machine intelligence},
  40(9):2109--2123, 2017.

\bibitem{zhou2017voxelnet}
Yin Zhou and Oncel Tuzel.
\newblock Voxelnet: End-to-end learning for point cloud based 3d object
  detection.
\newblock In {\em The IEEE Conference on Computer Vision and Pattern
  Recognition (CVPR)}, June 2018.

\end{thebibliography}
}

\newpage

\setcounter{section}{0}
\renewcommand\thesection{\Alph{section}} 
\section{Overview}
This document provides additional technical details and extra analysis experiments to the main paper.

In Sec.~\ref{sec:B}, we provide the information about the accuracy of mentioned depth maps while Sec.~\ref{sec:C} shows the detection performance using stereo images and LiDAR point clouds. 
Then, Sec.~\ref{sec:D} explains the correlation between 2D detector and resulting 3D detection performance.
Finally, Sec.~\ref{sec:E} presents detection results of {\it pedestrian} and {\it cyclist}.

\section{Accuracy of Depth Maps}
\label{sec:B}
Tab.~\ref{table:depth_mono} and Tab.~\ref{table:depth_stereo} show the accuracy of the monocular and stereo depth prediction methods listed in Tab.~\ref{table:depth}, respectively.
Combined with Tab.~\ref{table:depth}, it is evident that 3D detection accuracy increases significantly when using much more accurate depth (or disparity).
Note that the metrics for these two kinds of methods are different.

\begin{table}[!ht]
\begin{center}
\begin{tabular}{c||c|c|c|c}
\hline
~ & Abs Rel & Sq Rel & RMSE & RMSE$_{log}$ \\ 
\hline
MonoDepth & 0.097 & 0.896 & 5.093 & 0.176\\
DORN      & 0.071 & 0.268 & 2.271 & 0.116 \\
\hline
\end{tabular}
\end{center}
\caption{Accuracy of depth prediction (monocular) on KITTI {\it validation} set. lower is better.}
\label{table:depth_mono}
\end{table}

\begin{table}[!ht]
\begin{center}
\begin{tabular}{c||c|c|c}
\hline
~ & D1-bg & D1-fg & D1-all \\ 
\hline
DispNet & 4.32 \% & 4.41 \%  & 4.34 \%  \\
PSMNet  & 1.86 \% & 4.62 \%  & 2.32 \%  \\
\hline
\end{tabular}
\end{center}
\caption{Accuracy of depth prediction (stereo) on KITTI {\it test} set. lower is better.}
\label{table:depth_stereo}
\end{table}

\section{Extensions of Stereo and LiDAR}
\label{sec:C}
To further evaluate the proposed method, we extend it to stereo-based and LiDAR-based versions.
We select some representational methods based on stereo images  (or LiDAR point clouds) and report the comparative results in Table~\ref{table:extension}.
The experimental results show that our method is able to give a competitive performance when using LiDAR point clouds or stereo images as input.

Note that the proposed method with LiDAR point cloud input outperforms F-PointNet \cite{qi2017frustum} by {\bf 1.8} $AP_{3D}$, which proves that our RGB fusion module is equally effective for LiDAR-based methods.

\begin{table}[!ht]
\begin{center}
\begin{tabular}{c||c|c|c|c}
\hline
Method & Data & Easy & Mod. & Hard  \\ 
\hline
3DOP~\cite{chen20153d} & Stereo & 6.55 & 5.07 & 4.10  \\
Multi-Fusion~\cite{xu2018multi} & Stereo & - & 9.80 & -  \\
ours & Stereo & {\bf 45.85} & {\bf 26.03} & {\bf 23.16} \\ 
\hline
VoxelNet~\cite{zhou2017voxelnet} & LiDAR & 81.97 & 65.46 & 62.85  \\
FPointNet~\cite{qi2017frustum} & LiDAR & 83.26 & 69.28 & 62.56  \\
ours & LiDAR & {\bf 84.53} & {\bf 71.07} & {\bf 63.49}  \\
\hline
\end{tabular}
\end{center}
\caption{$AP_{3D}^{0.7}$(\%) of extended versions of proposed method and related works.}
\label{table:extension}
\end{table}

\section{2D Detectors}
\label{sec:D}
Tab.~\ref{table:2D} shows the correlation between the performance of 2D detectors and resulting 3D detection performance.
We can see that improving the performance of 2D detector is  an effective method to improve the overall detection accuracy.
However, the huge gap between the performance of 2D detector and final 3D estimator reveals there is still a lot of room for improvement without modifying the 2D detector.
The implementation details of the 2D detectors we used can be found in RRC~\cite{Ren_2017_CVPR} and F-PointNet~\cite{qi2017frustum}.

\begin{table}[!ht]
\begin{center}
\begin{tabular}{c|c|c|c|c|c|c}
\hline
~ & \multicolumn{3}{c|}{$AP_{2D}$} & \multicolumn{3}{c}{$AP_{3D}$} \\
\hline
~ & Easy & Mod. & Hard & Easy & Mod. & Hard \\ 
\hline
\cite{Ren_2017_CVPR}  & 88.4 & 86.7 & 76.6 & 31.1 & 20.0 & 16.8\\
\cite{qi2017frustum} & 90.5 & 89.9 & 80.7 & 32.2 & 21.1 & 17.3\\
\hline
\end{tabular}
\end{center}
\caption{Comparisons of different 2D detectors.
Metrics are $AP_{2D}$ and $AP_{3D}$ on KITTI {\it validation} set.}
\label{table:2D}
\end{table}

\section{Pedestrian and Cyclist}
\label{sec:E}
Most of previous image-based 3D detection methods only focus on {\it Car} category as KITTI provides enough instances to train their models.
Our model can also get a promising detection performance on {\it Pedestrian} and {\it Cyclist} categories because it is much easier and effective to do data augmentation for point clouds than depth maps used in previous methods. 
Table \ref{table:pedcyc} shows their $AP_{loc}$ and $AP_{3D}$ on KITTI {\it validation} set.

\begin{table}[!ht]
\begin{center}
\begin{tabular}{c|c|c|c|c|c}
\hline
Category & IoU & Task & Easy & Moderate & Hard \\ 
\hline
Pedestrian & 0.25 & Loc.  & 40.77 & 34.02 & 29.83 \\
Pedestrian & 0.25 & Det.  & 40.17 & 33.45 & 29.28 \\
Pedestrian & 0.5  & Loc.  & 14.30 & 11.26 & 9.23 \\
Pedestrian & 0.5  & Det.  & 11.29 &  9.01 & 7.04 \\
\hline
Cyclist & 0.25 & Loc.  & 28.15 & 17.79 & 16.57 \\
Cyclist & 0.25 & Det.  & 24.80 & 15.66 & 15.11 \\
Cyclist & 0.5  & Loc.  & 10.12 & 6.39 & 5.63 \\
Cyclist & 0.5  & Det.  & 8.90 & 4.81 & 4.52 \\
\hline
\end{tabular}
\end{center}
\caption{Benchmarks for Pedestrian and Cyclist. 3D localization and detection AP(\%) on KITTI {\it validation} set for {\it Pedestrian} and {\it Cyslist}. The IoU threshold is set to 0.25 and 0.5 for better comparison.}
\label{table:pedcyc}
\end{table}

\end{document}